\documentclass{article}

\PassOptionsToPackage{numbers, compress}{natbib}

    \usepackage[preprint]{neurips_2025}

\usepackage[utf8]{inputenc} 
\usepackage[T1]{fontenc}    
\usepackage{url}            
\usepackage{booktabs}       
\usepackage{amsfonts}       
\usepackage{xspace}
\usepackage{nicefrac}       
\usepackage{microtype}      
\usepackage{xcolor}         
\usepackage{xpatch}
\usepackage{microtype}
\usepackage{graphicx}
\usepackage{subfig}
\usepackage{booktabs} 
\usepackage[toc,page]{appendix}
\usepackage{titletoc}
\usepackage{listings}
\usepackage{algorithm}
\usepackage{algorithmic}
\usepackage{enumitem}
\usepackage{tabularx}
\usepackage{adjustbox}
\usepackage{longtable}
\usepackage{caption}
\usepackage{pifont}
\usepackage[utf8]{inputenc}
\usepackage{booktabs}
\usepackage{multirow}
\usepackage{amsmath, amssymb}
\usepackage{subcaption}
\usepackage{float}
\usepackage{rotating}
\usepackage{adjustbox}
\usepackage{microtype}
\usepackage{hyperref}
\usepackage{url}
\usepackage{booktabs}
\definecolor{darkblue}{rgb}{0, 0, 0.5}
\hypersetup{colorlinks=true, citecolor=darkblue, linkcolor=darkblue, urlcolor=darkblue}
\usepackage{xpatch}

\usepackage{microtype}

\usepackage{lineno}
\usepackage{graphicx}

\usepackage{amsmath}
\usepackage{wrapfig} 
\usepackage{lipsum}

\usepackage[most]{tcolorbox}
\usepackage{listings}
\tcbuselibrary{listings, breakable, skins}
\usepackage{tikz}
\usetikzlibrary{shadows}

\usepackage{listings}
 \usepackage{xcolor}

\usepackage{booktabs}
\usepackage{colortbl}
\usepackage{threeparttable}
\usepackage{array}
\usepackage{arydshln}

\usepackage[T1]{fontenc}

\usepackage{mathrsfs}
\usepackage{bm} 
\usepackage{graphicx}      
\usepackage{cleveref}

\usepackage{multirow, colortbl, caption}

\definecolor{lightblue}{RGB}{232, 244, 248}
\definecolor{lightpink}{RGB}{254, 238, 237}

\definecolor{bluelink}{RGB}{0,113,188}
\definecolor{greenlink}{RGB}{0,188,113}

\definecolor{darkblue}{rgb}{0, 0, 0.5}
\hypersetup{colorlinks=true, citecolor=darkblue, linkcolor=darkblue, urlcolor=darkblue}
\definecolor{darkyellow}{rgb}{0.980, 0.65, 0}

\definecolor{xolvergold}{rgb}{1.0, 0.75, 0.0} 

\definecolor{xolvercyan}{rgb}{0.0, 0.7, 0.9}
\definecolor{xolvergreen}{rgb}{0.0, 0.6, 0.3}

\newcommand{\xolver}{\textcolor{xolvergreen}{$\mathbb{X}$}olver\xspace}

\newcommand{\xolvercolor}{\textcolor{xolvergreen}{$\mathbb{X}$}olver\xspace}

\newcommand{\method}[1]{{\bfseries #1}}

\definecolor{headerLavender}{RGB}{230, 230, 250} 
\definecolor{rowLightGray}{RGB}{245, 245, 245} 
\definecolor{rowCream}{RGB}{255, 250, 240} 
\definecolor{errorRed}{RGB}{255, 77, 77} 

\definecolor{Red}{rgb}{1.0, 0.0, 0.0}
\definecolor{Yellow}{rgb}{1.0, 1.0, 0.0}
\definecolor{OliveGreen}{rgb}{0.33, 0.42, 0.18}
\definecolor{ForestGreen}{rgb}{0.13, 0.55, 0.13}

\usepackage{tikz}

\usepackage{tcolorbox}
\tcbset{highlight style/.style={
    colframe=gray!60,
    colback=gray!10,
    boxrule=0.3pt,
    arc=0.5mm,
    boxsep=0pt,
    left=1pt,
    right=1pt,
    top=1pt,
    bottom=1pt,
    enhanced,
    sharp corners}}
\newcommand{\highlightcell}[1]{\tcbox[highlight style]{#1}}

\makeatletter
\xapptocmd{\NAT@bibsetnum}{\setlength{\leftmargin}{0pt}\setlength{\itemindent}{\labelwidth}\addtolength{\itemindent}{\labelsep}}{}{}
\makeatother

\title{\xolver: Multi-Agent Reasoning with Holistic Experience Learning Just Like an Olympiad Team}

\author{%
Md Tanzib Hosain$^{1}$\thanks{Work done while working as a remote RA at QCRI.}  \quad
 Salman Rahman$^{2}$\footnotemark[1] \quad
Md Kishor Morol$^{3}$ \quad 
 Md Rizwan Parvez$^{4}$ \quad
 \\
 [1ex]
 $^{1}$American International University-Bangladesh \quad 
  $^{2}$ University of California, Los Angeles \quad  \\
 $^{3}$Cornell University \quad
 $^{4}$Qatar Computing Research Institute \quad
 \\
 \texttt{\href{mailto:mparvez@hbku.edu.qa}{mparvez@hbku.edu.qa}}
}

\begin{document}

\maketitle

\begin{abstract}
Despite impressive progress on complex reasoning, current large language models (LLMs) typically operate in isolation—treating each problem as an independent attempt, without accumulating or integrating experiential knowledge. In contrast, expert problem solvers—such as Olympiad or programming contest teams—leverage a rich tapestry of experiences: absorbing mentorship from coaches, developing intuition from past problems, leveraging knowledge of tool usage and library functionality, adapting strategies based on the expertise and experiences of peers, continuously refining their reasoning through trial and error, and learning from other related problems even during competition. Inspired by this, we introduce \method{\bf \xolvercolor}—a training-free, multi-agent reasoning framework that equips a black-box LLM with a persistent, evolving memory of holistic experience. 
\method{\xolver} integrates diverse experience modalities, including external and self-retrieval, tool use, collaborative agent interactions, agent-driven evaluation, and iterative reasoning refinement.
By learning from relevant strategies, code fragments, and abstract reasoning patterns at inference time, \method{\xolver} avoids generating solutions from scratch—marking a transition from isolated inference toward experience-aware language agents.
Built on both open-weight and proprietary models, \method{\xolver} consistently outperforms specialized reasoning agents (e.g., OctoTools, CheatSheet, Search-o1). Even when instantiated with lightweight backbones (e.g., QWQ-32B), it often surpasses the most advanced models to date—including Qwen3-235B, Gemini 2.5 Pro, o3, and o4-mini-high. With a stronger backbone like \texttt{o3-mini-high}, it achieves a new best result—98.1\% on GSM8K, 94.4\% on AIME'24, 93.7\% on AIME'25, 99.8\% on Math-500, and 91.6\% on LiveCodeBench—highlighting holistic experience learning as a key  step toward dynamic, generalist agents capable of expert-level reasoning. 
We open-source all code, and data of \xolver at \url{https://kagnlp.github.io/xolver.github.io/}. 
\begin{figure*}[h]
\centering
\vspace{-5pt}
\includegraphics[width=0.95\textwidth]{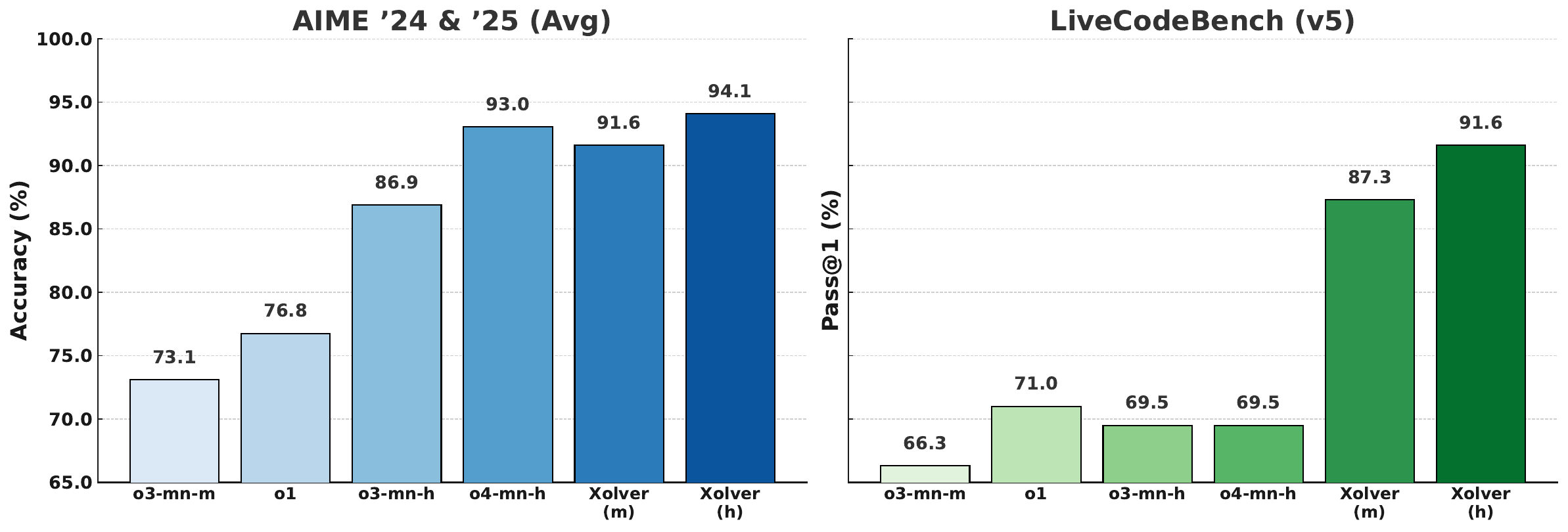}
\vspace{-5pt}
\caption{\textbf{Results Summary on  AIME '24 (16 runs), AIME '25 and LiveCodeBench (32 runs).} Our framework \method{\xolver}, built on \texttt{o3-mini-medium} and \texttt{o3-mini-high} backbones (denoted (m) and (h)), achieves up to 30.9\% gain over the baseline and often outperforms leading models on both tasks.}
 \vspace{-15pt}
\label{fig:splashfigure}
\end{figure*}

\end{abstract}

\section{Introduction}
\label{sec:introduction}

Recent advances in large language models (LLMs) have made remarkable progress in complex reasoning and problem solving across domains such as mathematics~\citep{cobbe2021gsm8k, hendrycks2021math, lewkowycz2022solving} and programming \citep{chen2021humaneval, austin2021program, khan-etal-2024-xcodeeval}. Yet despite these impressive capabilities, conventional LLM reasoning approaches remain fundamentally limited: they standalone each problem instance, generating solutions from scratch without accumulating or transferring insights from rich, diverse experiential knowledge.

This isolated reasoning paradigm marks a significant departure from how expert human problem solvers operate. Expert problem solvers—such as an Olympiad or programming contest teams—rarely approach problems in a vacuum. Instead, they draw upon a rich tapestry of cumulative experiences: absorbing mentorship from coaches, developing intuition from past problems, leveraging knowledge of tool usage and library functionality (e.g., calculator), adapting strategies based on peers’ expertise and experiences, gaining insights through iterative trial and error, and learning from related problems even during competition. This holistic experience empowers them to tackle new challenges not from scratch, but by dynamically applying accumulated knowledge and adaptive strategies.

While numerous prior studies have enhanced LLM reasoning and problem solving through various forms of experiential knowledge augmentation, they have predominantly operated within discrete modalities—retrieving similar problems or relevant contexts \citep{parvez-etal-2021-retrieval-augmented, lewis2020retrieval, guu2020realm}, leveraging external tools \citep{lu2025octotools, lu2024chameleon}, or facilitating multi-agent collaboration \citep{islam-etal-2025-codesim, hong2023metagpt, wu2023autogen}. Despite their individual strengths, these approaches address distinct facets of experiential knowledge independently, preventing LLMs from accumulating and synthesizing a comprehensive repertoire of learning signals across diverse experiential dimensions, thereby limiting the development of the rich, interconnected knowledge structures that characterize human expertise.

In this paper, we introduce \method{\xolver}, a unified, memory-augmented, multi-agent inference framework that emulates the holistic experience-driven, collaborative reasoning of expert teams.
\xolver dynamically orchestrates a roster of specialized agents—such as mathematicians, programmers, verifiers—that iteratively tackle complex problems. Unlike conventional LLM pipelines, \xolver seamlessly integrates planning, episodic retrieval—both from external or self-parametric long-term memory—an evolving intermediate shared memory, tool invocation, multi-agent collaboration, agent-driven evaluation, and iterative self-refinement into a single adaptive architecture.

\begin{figure*}[h]
\centering
\vspace{-10pt}
\includegraphics[width=\textwidth]{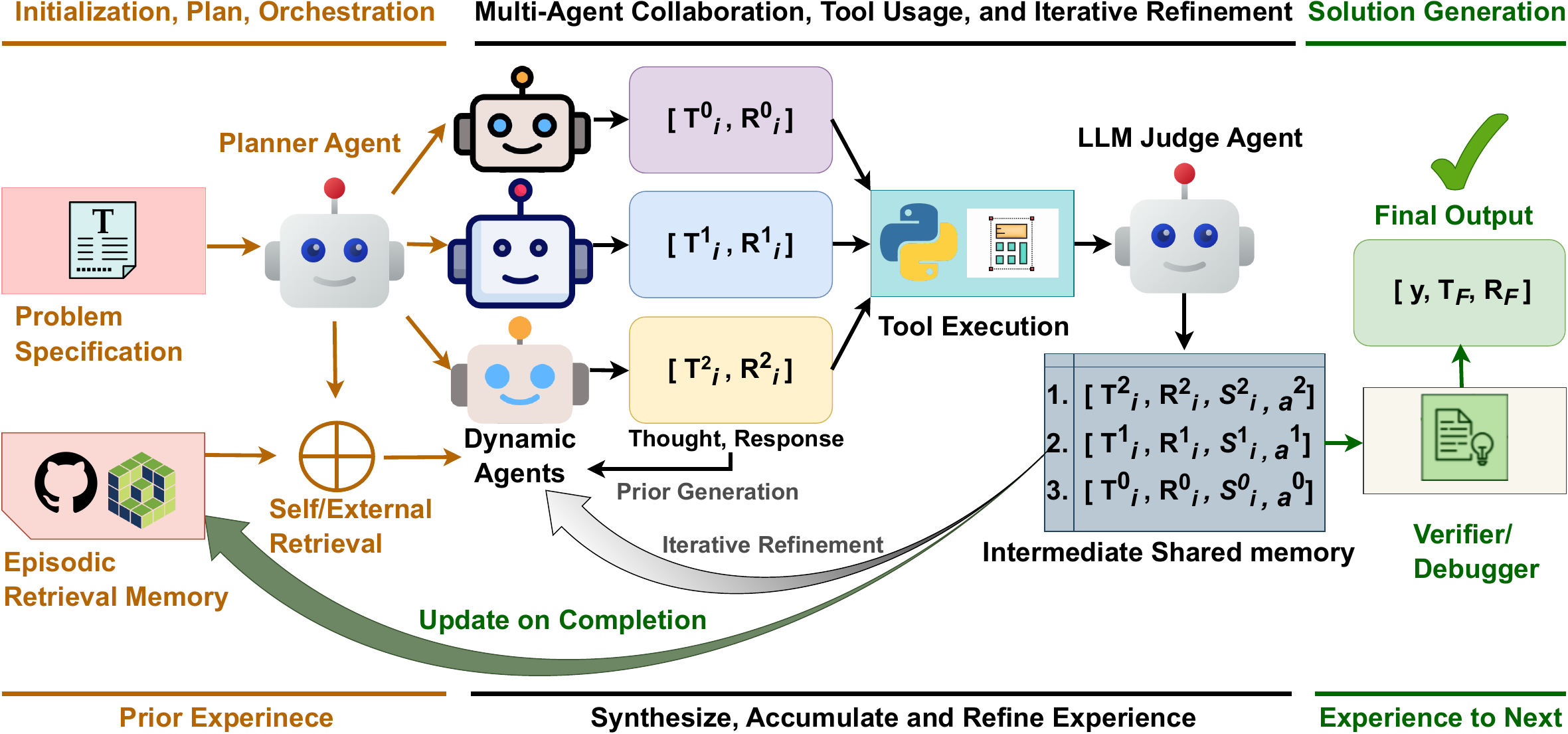}
\caption{\textbf{\method{\xolver} Scaffold.} At each iteration, agents receive their past reasoning history and top-ranked exemplars to generate new thoughts and responses, using tools (e.g., code) as needed. A judge model ranks outputs, and an intermediate memory maintains the best responses over time. Exemplars are initialized via episodic retrieval and continually updated with high-quality solutions from the memory. Iteration stops when convergence or max steps are reached, followed by final verification.
}
\vspace{-15pt}
\label{fig:bannerfigure}
\end{figure*}

Each agent’s reasoning begins with exemplars drawn from episodic memory. From the second iteration onward, agents rely exclusively on an evolving shared memory that records the highest-quality reasoning paths, solutions, and evaluation feedback generated so far—thereby accumulating symbolic experience over time. This shared repository guides agents to build on successful strategies, correct mistakes, and improve solution quality. When needed, agents invoke external tools (e.g., code execution), and a dedicated judge agent reviews all outputs—selecting top responses, issuing feedback, and enriching the intermediate shared memory with curated traces and collective evaluations for future rounds. Iterations continue until outputs converge or a preset limit is reached, followed by a final verification or external debugging phase to ensure correctness. Additionally, by updating its episodic store with each newly solved problem and its reasoning trace, \xolver can continually expand its knowledge base. Through this closed loop of collaborative agents, memory-guided refinement, and tool guided precision, \xolver features a more holistic experience learning and transcends static LLM inference, delivering adaptive, expert-level reasoning over time. Figure \ref{fig:bannerfigure} illustrates the workflow.

We conduct large-scale experiments across a range of math and programming benchmarks—including GSM8K, Math-500, AIME (2024 and 2025), and LiveCodeBench (v5)—using both proprietary ({o3-mini-medium}) and open-weight ({QWQ-32B}) backbone models. \method{\xolver} consistently outperforms specialized reasoning systems such as OctoTools~\cite{lu2025octotools}, CheatSheet~\cite{suzgun2025dynamic}, and Search-o1~\cite{li2025search}. Remarkably, even when instantiated with lightweight models, \xolver often surpasses significantly larger state-of-the-art LLMs, including Qwen3-235B \citep{qwen3}, Gemini 2.5 Pro \citep{gemini2.5}, o1, o3, and o4-mini-high~\cite{openai2024o1}. As in \Cref{fig:splashfigure}, \xolver (m) achieves 91.6\% average accuracy on the AIME~'24 and~'25 benchmarks—an 18.5-point gain over {o3-mini-medium}—while \xolver (h) reaches 94.1\%, outperforming {o3-mini-high} by 7.2 points. On {LiveCodeBench}, \xolver (m) improves upon its base by 21 points (66.3\% to 87.3\%), with \xolver (h) achieving 91.6\%, a 22.1-point lift over {o3-mini-high}.

Our analysis reveals how \xolver's experiential components contribute to its performance. Accuracy improves consistently with more agents and iterations, reflecting the benefits of experience accumulation, though at increased cost. While external retrieval remains powerful, we find that self-retrieval—drawing from the model’s own parametric memory—can serve as an alternative with some performance drop. For tasks involving symbolic reasoning and complex arithmetic, multi-agent, multi-iterative refinement is more beneficial than tool use (e.g., Python execution). Our experiments confirm that even without updating episodic memory during inference, \xolver retains substantial performance gains, emphasizing the strength of its intermediate memory and iterative refinement. Together, these findings highlight \xolver’s ability to accumulate, refine, and reuse symbolic experience through collaborative, memory-guided reasoning. 

\section{The \xolver Framework}
\label{sec:system}

Given a problem query \( q \in \mathcal{Q} \) and a pretrained language model \( \mathrm{LLM}_\theta(\cdot) \), a conventional approach generates a solution via single-step inference: \( y \sim \mathrm{LLM}_\theta(q) \). In contrast, \method{\xolver} executes a dynamic, multi-agent reasoning process that iteratively accumulates and leverages symbolic experience to solve complex problems more effectively.

To support structured collaborative reasoning, \xolver maintains two complementary forms of memory: an \textit{episodic memory} \( \mathcal{D}_E \), which stores a library of past problems, solutions, and reasoning traces; and an intermediate dynamic \textit{shared memory} \( \mathcal{D}_S \), which evolves during inference to retain high-quality agent trajectories—comprising reasoning thoughts, responses, agent metadata, and feedback. In \xolver, a multi-agent team \( \mathcal{A} \) is orchestrated adaptively by a planner agent \( \mathcal{P} \), which assigns roles and configures memory access. During inference, \( \mathcal{A} \) agents leverage an external toolset \( \mathcal{T} \) (e.g., Python interpreter) to support accurate computation. Finally, a verifier or external debugger \( \mathcal{V} \) is invoked to extract and format the final answer, and to validate correctness for executable outputs.

Below, we first describe the \xolver agents and tools in Section \ref{sec:method:agents}, followed by the memory components in Section \ref{sec:method:memory}, and the inference cycle in Section \ref{sec:method:inference}.

\subsection{Agents and Tools}
\label{sec:method:agents}

\noindent{\bf Planner Agent \( \mathcal{P} \).}
The planner agent \( \mathcal{P} \) is responsible for initiating, planning, and orchestrating the \xolver multi-agent architecture. Given the problem \( q \) and the number of agents $m$, it constructs a team \( \mathcal{A} \) of \( m \) dynamic agents, each assigned a distinct expert role (e.g., algebra solver, mathematician, theorist, programmer, algorithm designer) tailored to the demands of \( q \). To ensure sufficient task coverage and role diversity, \( \mathcal{P} \) first prompts the underlying LLM to over-generate \( M > m \) candidate agents, from which it then selects the most effective subset \( \mathcal{A} \subset \{a_1, \ldots, a_M\} \) such that \( |\mathcal{A}| = m \). A summary of the most frequently generated and selected roles is provided in Appendix \ref{sec:appendix:role-list}.

\noindent{\bf Dynamic Reasoning Agents \( \mathcal{A} \).}
The set \( \mathcal{A} = \{a^1, a^2, \ldots, a^m\} \) represents a team of dynamic reasoning agents constructed by the planner agent \( \mathcal{P} \). Each agent \( a^j \in \mathcal{A} \) is assigned a distinct expert role (e.g., algebra solver, programmer, counter-example generator) tailored to the task query \( q \). Agents are instantiated using a standardized prompting template (see \Cref{sec:appednix:prompts}) that incorporates the task description, assigned role, retrieved examples, prior reasoning attempts, and shared memory feedback—enabling iterative self-correction and role specialization.

At each iteration \( i \), agent \( a^j \) receives a context \( \mathcal{C}_i^j \) and generates a structured reasoning trace \( T_i^j \) and a response \( R_i^j \).  For the first iteration (\( i = 0 \)), the context is initialized using the task query and relevant retrieved exemplars:
\[
\mathcal{A} \leftarrow \mathcal{C}_0^j = \{q\} \cup \mathcal{R}(\mathcal{D}_E) \tag{\textsc{BuildContext}}.
\]
For subsequent iterations (\( i \geq 1 \)), the context evolves by incorporating its prior generation (history) and the shared memory: 
\[
\mathcal{A} \leftarrow \mathcal{C}_i^j = \{q\} \cup \{ T_{i-1}^j, R_{i-1}^j \} \cup \mathcal{D}_S \tag{\textsc{BuildContext}}.
\]

\paragraph{\bf Judge Agent \( \mathcal{J} \).}
The judge agent \( \mathcal{J} \) evaluates intermediate outputs from each agent and returns structured feedback to guide refinement and memory updates. Given a query \( q \), a reasoning trace \( T \), and a response \( R \), it produces a feedback tuple \( S = (T_S, s) \), where \( T_S \) is a natural language explanation (e.g., critique, justification, correction), and \( s \) is a scalar quality score. The interpretation of \( s \) is task-dependent: for math problems, \( s \in [0, 1] \) reflects an LLM-estimated correctness probability; for code tasks, \( s \in \{0, 1, \dots, N_{\text{test}}\} \), where \( N_{\text{test}} \) denotes the total number of test cases including problem-provided samples and 10 synthesized test cases generated using AceCode-RM-32B~\citep{zeng2025acecoder}. To avoid compiler interaction latency and maintain symbolic traceability, test case outcomes are determined by simulating execution through LLM prompting within the judge agent \( \mathcal{J} \), following the CodeSim protocol \citep{islam-etal-2025-codesim}. This structured feedback enables agents to identify failures, receive localized corrections, and improve reasoning over iterations.

\paragraph{\bf Verifier Agent \( \mathcal{V} \).}
Due to linguistic complexity and varying answer specification formats, a response may be incorrect even when the underlying reasoning or open-ended response is valid. For instance, answer formats may require multiple-choice letters (e.g., “(A)” or “Choice B”), boxed numerical values (e.g., “\boxed{42}”), or final answers in specific units (e.g., “5 km” or “12\%”). An additional round of answer extraction and formatting helps reduce such mispredictions~\citep{parvez-2025-chain}. This challenge is even more pronounced in code generation tasks, where predicted code may fail to execute or not pass all test cases. To mitigate this, \xolver includes a Verifier Agent \( \mathcal{V} \), which operates differently based on the output type. For math and QA problems, \( \mathcal{V} \) extracts the final reasoning \( T_F \), response \( R_F \), and answer \( y \) from the response associated with the top-ranked entry $\textsc{BestResponse}$ in \( \mathcal{D}_S \), ensuring adherence to the expected output format. For executable code, \xolver invokes an external debugger (LDB~\citep{zhong2024ldb}), where \( \mathcal{V} \) interacts with a Python runtime to capture execution feedback and iteratively fix runtime errors. 

\paragraph{\bf Tools \( \mathcal{T} \).}
Integrating natural language reasoning with tools like Python execution is a proven way to boost performance on complex reasoning tasks~\cite{openmathreason, toshniwal2024openmathinstruct}. We observe that even advanced reasoning models often make mistakes in intermediate steps, particularly when computations become non-trivial. To address this, each dynamic agent \( a^j \) is explicitly instructed to use Python execution during reasoning when needed. While \xolver currently limits \( \mathcal{T} \) to Python, our prompting strategy is tool-agnostic, allowing an interface for future extensions to richer toolsets \citep{lu2024chameleon,lu2025octotools}.

All agents are built using the underlying LLM. All prompts are 0-shot and provided in Appendix \ref{sec:appednix:prompts}.


\subsection{Memory Components}
\label{sec:method:memory}


\noindent{\bf Episodic Memory \( \mathcal{D}_E \).} 
 \xolver maintains two forms of episodic (long-term) memory: (1) an external memory corpus \(\mathcal{D}_E^{\text{ext}} = \{(q', T', R')\}\), which consists of past problem instances \(q'\), their corresponding reasoning traces \(T'\) (optional), and solution responses \(R'\); and (2) the internal parametric memory encoded in the weights of the agent-specific language model \(\mathrm{LLM}_{j}\).

We define a general retrieval operator \(\mathcal{R}(\mathcal{D}_E)\) that returns a set of \(K\) examples relevant to the query \(q\). When \(\mathcal{D}_E^{\text{ext}}\) is available, retrieval is conducted using similarity-based search (e.g., BM25):
\[
\mathcal{R}(\mathcal{D}_E) = \{(q'_k, T'_k, R'_k)\}_{k=1}^{K} \leftarrow \mathrm{Retrieve}_j(q, \mathcal{D}_E^{\text{ext}}).
\]
Otherwise, \xolver falls back to internal self-retrieval by sampling from the agent model itself:
\[
\mathcal{R}(\mathcal{D}_E) = \{(q'_k, T'_k, R'_k)\}_{k=1}^{K} \sim \mathrm{LLM}_{j}(q).
\]
In the case of an external episodic memory, $\mathcal{D}_E$ can also be updated with $\textsc{UpdateEpisodicMemory}$ by adding the top-ranked reasoning and response from $\mathcal{D}_S$, paired with the problem $q$, into the external corpus \(\mathcal{D}_E^{\text{ext}}\). That is,
\(
\mathcal{D}_E^{\text{ext}} \leftarrow \mathcal{D}_E^{\text{ext}} \cup (q, T, R),
\)
where $(T, R, S, a)$ is the top-ranked entry in $\mathcal{D}_S$.

\paragraph{\bf Intermediate Shared Memory $\mathcal{D}_S$.}
The shared memory $\mathcal{D}_S$ maintains a fixed-size set of high-quality intermediate reasoning, responses, and metadata generated by the dynamic agents during inference on the current query $q$. For simplicity and to preserve the dynamic nature of the framework, we constrain $|\mathcal{D}_S| = m$, where $m$ is the number of dynamic agents in $\mathcal{A}$. Initially, $\mathcal{D}_S \leftarrow \emptyset$. At each iteration $i$, each agent $a_j \in \mathcal{A}$ produces a reasoning trace $T_i^j$, response $R_i^j$, and receives structured feedback $S_i^j = (T_S^{(i,j)}, s_{i,j})$ from the judge agent $\mathcal{J}$, where $T_S^{(i,j)}$ is a natural language explanation and $s_{i,j}$ is a scalar score reflecting the quality of the tuple $(T_i^j, R_i^j)$. After collecting the new outputs 
\[
\tau_i^j = (T_i^j, R_i^j, S_i^j, a^j),\quad j=1,\dots,m, \tag{\textsc{RunAgents}}
\] we form the candidate pool
\(
\mathcal{M} = \mathcal{D}_S \cup \{\tau_i^1, \dots, \tau_i^m\}. 
\)
We then update the fixed-size shared memory by keeping only the top-$m$ tuples by score
\[
\mathcal{D}_S \leftarrow \mathrm{TopK}\bigl(\mathcal{M},\; m;\; \mathrm{key}(e) = s(e)\bigr), \tag{\textsc{UpdateSharedMemory}},
\]
where \(s(e)\) extracts the scalar score from \(e=(T,R,(T_S,s), a)\).  

This replacement mechanism ensures that $\mathcal{D}_S$ always contains exactly $m$ entries with the highest observed scores across all iterations. By maintaining only the strongest reasoning-response-feedback tuples, the shared memory facilitates knowledge transfer between agents and across iterations, enabling collaborative improvement through exposure to diverse high-quality solutions.

\begin{wrapfigure}{r}{0.55\textwidth}
\vspace{-4em}
\begin{minipage}{0.55\textwidth}
\begin{algorithm}[H]
\captionof{algorithm}{\xolver Inference Protocol}
\label{algo:xolver}
\begin{algorithmic}[1]
\STATE  \textbf{Input:} Query $q$, Tools $\mathcal{T}$, Episodic Memory $\mathcal{D}_E$, parameters \( m, k, I \)
\STATE  \textbf{Init:} $\mathcal{D}_S \gets \emptyset$
\STATE  $\mathcal{A} \gets \textsc{Planner}(q, m)$
\FOR{$i = 0$ \textbf{to} $\mathcal{I}$}
    \STATE  $\{\mathcal{C}_i\}_{c=1}^m \gets \textsc{BuildContext}(\mathcal{A}, \mathcal{D}_E, \mathcal{D}_S, q, i)$
    \STATE  $\{\tau_i^j\}_{j=1}^m \gets \textsc{RunAgents}(\mathcal{A}, \mathcal{C}_i, \mathcal{T}, \mathcal{J})$
    \STATE  $\mathcal{D}_S \gets \textsc{UpdateSharedMemory}(\mathcal{D}_S, \{\tau_i^j\})$
    \IF{\textsc{Converged}$(\mathcal{D}_S)$} \STATE \textbf{break} \ENDIF
\ENDFOR
\STATE $y \gets \mathcal{V}(\textsc{BestResponse}(\mathcal{D}_S))$
\STATE \textsc{UpdateEpisodicMemory}$(\mathcal{D}_E, q, \mathcal{D}_S)$
\STATE \textbf{Return} $y$

\end{algorithmic}
\end{algorithm}
\end{minipage}
\vspace{-2em}
\end{wrapfigure}

\subsection{Inference Protocol}
\label{sec:method:inference}

Algorithm \ref{algo:xolver} summarizes the \xolver inference protocol, which operates in three structured stages. 
\textbf{Stage-1}, which emulates initialization with prior experience, involves the planner constructing a team of agents \( \mathcal{A} \) (lines 2--3). 
\textbf{Stage-2}, embodying symbolic experience accumulation and refinement, iterates for \( \mathcal{I} \) rounds (lines 4--10). 
In each round, all agents receive access to \( \mathcal{D}_S \) and \( \mathcal{D}_E \), build their contexts, and generate structured trajectories and responses 
(\( \mathcal{D}_E \) is only used for context construction at the first iteration). 
These are evaluated by the judge agent \( \mathcal{J} \), and \( \mathcal{D}_S \) is updated with the resulting feedback tuples (line 7). 
Upon convergence or after \(\mathcal{I}\) rounds, \textbf{Stage-3} invokes the verifier agent \( \mathcal{V} \), which extracts the final answer from the top-ranked entry in \( \mathcal{D}_S \) (line 11), and updates \( \mathcal{D}_E \) with the new experience.

\section{Experiments}
\label{sec:exp}

\subsection{Evaluation Setup}
\label{sec:exp:setups}

\noindent {\bf Evaluation Benchmarks}
We evaluate \xolver across five diverse and challenging benchmarks covering both mathematical and coding reasoning. For math, we use GSM8K~\citep{cobbe2021gsm8k}, Math-500\citep{hendrycks2021math}, and the AIME 2024  \citep{AIME2024} and  2025 \citep{AIME2025}, comprising high-school level competition problems requiring multi-step symbolic reasoning. For coding, we use LiveCodeBench~(v5)~\citep{jain2024livecodebench}, a dynamic benchmark that ensures no data leakage by periodically releasing new problems. These benchmarks span arithmetic, algebra, number theory, geometry, combinatorics, and algorithmic problem solving.

\noindent {\bf Baselines and Metrics}
We compare \xolver against directly using leading reasoning models--
\textit{(a) proprietary models}: 
Gemini~2.5 (Pro and Flash Think) ~\citep{gemini2.5}, 
Grok-3 Beta Think and Grok-3 Mini (Beta) Think~\citep{grok3}, 
Claude~3.7 Sonnet Think~\citep{claude3.7}, 
{o1}~\citep{openai2024o1}, 
{o3-mini}, {o3}, and {o4-mini}~\citep{o3o4}; 
\textit{(b) open-weight LLMs}, 
e.g., Qwen3-235B~\citep{qwen2024qwq}, 
QWQ-32B~\citep{qwq32b}, 
and DeepSeek-R1~\citep{deepseekai2025r1}; 
\textit{(c) math- and code-specialized models}, 
e.g., AlphaOne~\citep{alphaone}, OpenMathReason~\citep{openmathreason}, 
rStar-Math~\citep{guan2025rstarmathsmallllmsmaster}, rStar-Coder~\citep{rstar-coder}, 
OpenCodeReason~\citep{ahmad2025opencodereasoningadvancingdatadistillation}, 
and Kimi-K1-1.6~\citep{kimi2025k15}. We also compare with
\textit{(d) agents or frameworks}:
Self-Reflexion~\citep{shinn2023reflexion}, 
agentic search based framework Search-o1~\citep{li2025search}, 
specialized tool based framework OctoTools~\citep{lu2025octotools} which excels general purpose agent platforms outperforming AutoGen or LangChain, 
cross-problem baseline framework CheatSheet~\citep{suzgun2025dynamic}, 
and multi-agent code generation framework CodeSim~\citep{islam-etal-2025-codesim}, 
which leverage refinement, retrieval or online search, fine-grained tool augmentation in addition to online search, dynamic memory updates after solving new problems, and multi-agent reasoning techniques respectively. For agent-based baselines (d), we reproduce results using the same backbone LLMs as \xolver for fair comparison; for model-based baselines (a–c), we report official results from their technical reports or corresponding benchmark leaderboards. 
As evaluation metric, we use accuracy using GPT-4o~\citep{gpt4o} for math problems, and \textit{pass@1}  for code tasks.

\noindent {\bf Inference Details}
We use both open-weight {QWQ-32B}~\citep{qwen2024qwq} and proprietary {o3-mini} (medium and high)~\citep{o3o4} as the backbone. To mitigate performance variance inherent in single-run evaluations, we report the average accuracy and \textit{pass@1} metric, calculated by averaging 32 inference runs for competitive benchmarks \textsc{LiveCodeBench} and \textsc{AIME '25}, and 16 runs for \textsc{AIME '24}, ensuring standard deviation within $\sim1\%$ (Appendix \ref{sec:appendix:reasoning}). 
For simpler tasks \textsc{GSM8K} and \textsc{Math-500}, we follow DeepSeek-v3~\citep{liu2024deepseek}, using a single greedy-decoded generation. By default, we set temperature to 0.2, number of agents $m=3$, and max iterations $\mathcal{I}=2$. \xolver iteration terminates either when the maximum number of iterations $\mathcal{I}$ is reached, or when all entries in the shared memory $\mathcal{D}_s$ converge—i.e., they achieve perfect scores of 1.0 (correct) for math tasks, or pass all test cases (both sample and synthesized) for code tasks.
As the external retrieval corpus $\mathcal{D}_E^{\text{ext}}$ in coding task, we collect a 9-million-token dataset of algorithmic code problems and their C++ solutions with explanations from GitHub\footnote{\url{https://github.com/cp-algorithms/cp-algorithms}} (details in Appendix \ref{sec:appendix:ext_retrieve}). For math, we use the \textsc{OpenMathReason} dataset~\citep{openmathreason} as $\mathcal{D}_E^{\text{ext}}$.
We evaluate two variants of \method{\xolver}: (i) \textit{\xolver with in-competition cross-problem experience} (\xolver (+)), which dynamically updates the episodic memory after solving each problem to utilize accumulated knowledge across problems; and (ii) \textit{\xolver (–)}, which keeps the episodic memory static, focusing solely on  problem-specific experience. By default, we refer to \xolver (+) as our method if not specified otherwise.

\subsection{Main Results}
\label{sec:exp:main-results}

\begin{table}[t]
\centering
\resizebox{0.9\textwidth}{!}{%
\begin{tabular}{@{}llccccc@{}}
\toprule
\textbf{Model} & \textbf{Appr.}
& \textbf{GSM8K}
& \shortstack{\textbf{AIME}\\\textbf{'24}}
& \shortstack{\textbf{AIME}\\\textbf{'25}}
& \shortstack{\textbf{Math}\\\textbf{-500}}
& \shortstack{\textbf{LiveCodeBench}\\\textbf{(v5)}} \\
\midrule
\multicolumn{7}{c}{\textbf{Proprietary Models}} \\
\midrule
Claude 3.7 Sonnet T.      & LongCoT & –    & 61.3 & 49.5 & 96.2 & 51.4 \\
Grok-3 (Beta) T.      & Direct & –    & 83.9 & 77.3 & –    & 70.6 \\
Grok-3-mini (Beta) T. & LongCoT & –    & 89.5 & 82.0 & –    & - \\
Gemini 2.5 Flash T.   & LongCoT & –    & 88.0 & 78.0 & –    & 63.5 \\
o1                   & LongCoT & 96.4 & 74.3 & 79.2 & 96.4 & 71.0 \\
o3-mini-high         & LongCoT & –    & 87.3 & 86.5 & –    & 69.5 \\
Gemini 2.5 Pro.      & Direct & –    & 92.0 & 86.7 & –    & 70.4 \\
o3                   & LongCoT & 96.7 & 91.6 & 88.9 & –    & –    \\

o4-mini-high         & LongCoT & –    & 93.4 & \textcolor{black}{\textbf{92.7}} & – & 69.5 \\
\midrule
\multicolumn{7}{c}{\textbf{Open Weights Models}} \\
\midrule
DeepSeek-R1          & LongCoT & –    & 79.8 & 70.0 & 97.3 & 64.3 \\
Qwen3-235B-A22B      & LongCoT & –    & 85.7 & 81.5 & –    & 70.7 \\
\midrule
\multicolumn{7}{c}{\textbf{Math/Code Specialized Models}} \\
\midrule
\rowcolor{rowLightGray}
rStar-Math (Best)    & – & 95.2 & 53.3 & –    & 90.0 & –    \\
\rowcolor{rowLightGray}
OpenMathReason (Best)      & – & –    & {93.3} & 80.0 & –    & –    \\
\rowcolor{gray!2}
AlphaOne (Best) & – & – & 53.3 & – & 89.4 & 75.8 \\
\rowcolor{gray!15}
OpenCodeReason (Best)           & – & –    & –    & –    & –    & {61.8} \\
\rowcolor{gray!15}
rStar-Coder  (Best)           & – & –    & –    & –    & –    & {62.5} \\
\rowcolor{gray!15}
Kimi-k1.6-IOI-high   & – & –    & –    & –    & –    & {73.8} \\
\midrule
\multicolumn{7}{c}{\textbf{Reasoning Agents/Frameworks}} \\
\midrule
\multirow{6}{*}{o3-mini-medium}
  & LongCoT       & 95.2 & 75.8 & 70.4 & 97.3 & 66.3 \\
  & Self-Refl.   & 93.1 & 79.4 & 76.5 & 95.2 & 73.2 \\
  & OctoTools    & 95.4 & 81.7 & 75.3 & 97.5 & –    \\
  & Search-o1 &  95.8 & 81.8 & 76.7 & 97.6 & 73.6 \\
  & CheatSheet & 95.9 & 82.2 & 75.8 & 97.7 & – \\
  & CodeSim & – & – & – & – & 73.8 \\
  \rowcolor{blue!10}
  & \method{\bf \xolver} (–) & 95.6 & 87.2 & {85.1} & \underline{97.7} & 79.6 \\
  \rowcolor{blue!10}
  & \method{\bf \xolver} (+) & \underline{97.1} & \textbf{93.8} & \underline{89.4} & \textbf{99.2} & \textbf{87.3} \\
\midrule
\multirow{7}{*}{QWQ-32B}
  & LongCoT       & 96.1 & 78.1 & 65.8 & 83.2 & 63.4 \\
  & Self-Refl.   & 94.0 & 79.3 & 66.3 & 80.4 & 69.2 \\
  & OctoTools    & 96.3 & 83.0 & 71.7 & 86.1 & –    \\
  & Search-o1 & 96.4  & 84.4 & 71.8 & 87.1 & 69.3 \\
  & CheatSheet & 96.8 & 83.5 & 72.2 & 86.5 & –   \\
  & CodeSim & – & – & – & – & 70.5 \\
  \rowcolor{blue!10}
  & \method{\bf \xolver} (–) & 96.5 & 89.9 & 79.5 & 93.1 & 76.2 \\
  \rowcolor{blue!10}
  & \method{\bf \xolver} (+) & \textbf{98.0} & \underline{93.6} & {82.7} & 95.5 & \underline{79.2} \\
  \midrule
  \rowcolor{green!10}
        o3-mini-high & \method{\bf \xolver (+) } & \textbf{98.1} & \textbf{94.4} & \textbf{93.7} & \textbf{99.8} & \textbf{91.6} \\

\bottomrule
\end{tabular}
}
\vspace{1pt}
\caption{Comparison of \method{\xolver} against SoTA reasoning models, specialized models, and other reasoning agents across mathematical and coding tasks. Best results are boldfaced and second-best results are underlined. T: Think models, LongCoT*: standard prompting for reasoning models. "-" denotes either n/a (e.g., only math/code specialized models) or results not reported.  
\vspace{-22pt}
}

\label{tab:merged_reasoning_performance}
\end{table}





Table \ref{tab:merged_reasoning_performance} evaluates \xolver across diverse mathematical and coding reasoning benchmarks, highlighting its effectiveness compared to state-of-the-art LLMs, specialized models, and other frameworks.

\noindent {\bf Strong Gains Across Benchmarks}  
Overall, \xolver consistently delivers significant improvements over the backbone LLMs’ standard LongCoT prompting. Both the problem-specific \xolver (–) and the cross-problem \xolver (+) variants outperform their respective backbone LLM (LongCoT) baselines across all datasets. For example, with the o3-mini-medium backbone, \xolver (+) improves from 75.8 to 93.8 on AIME’24, and from 66.3 to 79.6 on LiveCodeBench, while the QWQ-32B backbone sees gains from 78.1 to 89.9 on AIME’24 and from 63.4 to 76.2 on LiveCodeBench. 

\noindent {\bf  Surpassing Prior Agents}  
Compared to previous frameworks such as Search-o1, OctoTools, and CheatSheet, \xolver demonstrates consistent and significant gains. With o3-mini-medium, \xolver (+) improves over the best baseline by +12.7 points on AIME'25 and +13.5 points on LiveCodeBench, highlighting its superior reasoning capabilities by integrating diverse forms of experience.

\noindent {\bf In Comparison to Leading LLMs}  
Despite using weaker backbones, \method{\xolver}, specifically (+) variant, matches or surpasses proprietary frontier LLMs like o3 and o4-mini-high on key benchmarks. With o3-mini-medium,  \method{\xolver (+} outperforms o4-mini-high on AIME'24 (93.8 vs. 93.4) and substantially exceeds it on LiveCodeBench (87.3 vs. 69.5), demonstrating that structured reasoning and dynamic memory can rival even the strongest closed-source models.

\noindent {\bf Backbone Agnostic}  
Improvements from \xolver are consistent across different backbone LLMs. Both o3-mini-medium and QWQ-32B benefit substantially from the framework, demonstrating its model-agnostic design. For example, on GSM8K, \xolver (+) achieves 97.1 (o3-mini-medium) and 98.0 (QWQ-32B), both surpassing baseline variants by significant margins.

\noindent {\bf Effectiveness of Dynamic Episodic Memory}  
While both variants excel, the cross-problem variant \xolver (+) consistently outperforms the problem-specific version \xolver (-) in all benchmarks. On average, episodic memory integration yields a +3.5 point improvement across both backbones and datasets where the largest gain is +7.7 points with o3-mini-medium on coding (LiveCodeBench).

\noindent\textbf{Scales with Backbone LLM's Strength}  
\method{\xolver}'s performance scales consistently with the strength of its backbone LLM. With {o3-mini-high}, it sets new state-of-the-art results across all benchmarks (98.1 on GSM8K, 94.4 on AIME'24, 93.7 on AIME'25, 99.8 on Math-500, and 91.6 on LiveCodeBench).

\section{Ablation and Analyses}
\label{sec:ablation-analysis}
\begin{wrapfigure}{r}{0.6\textwidth}
    \vspace{-5pt}
    \centering
    \includegraphics[width=\linewidth]{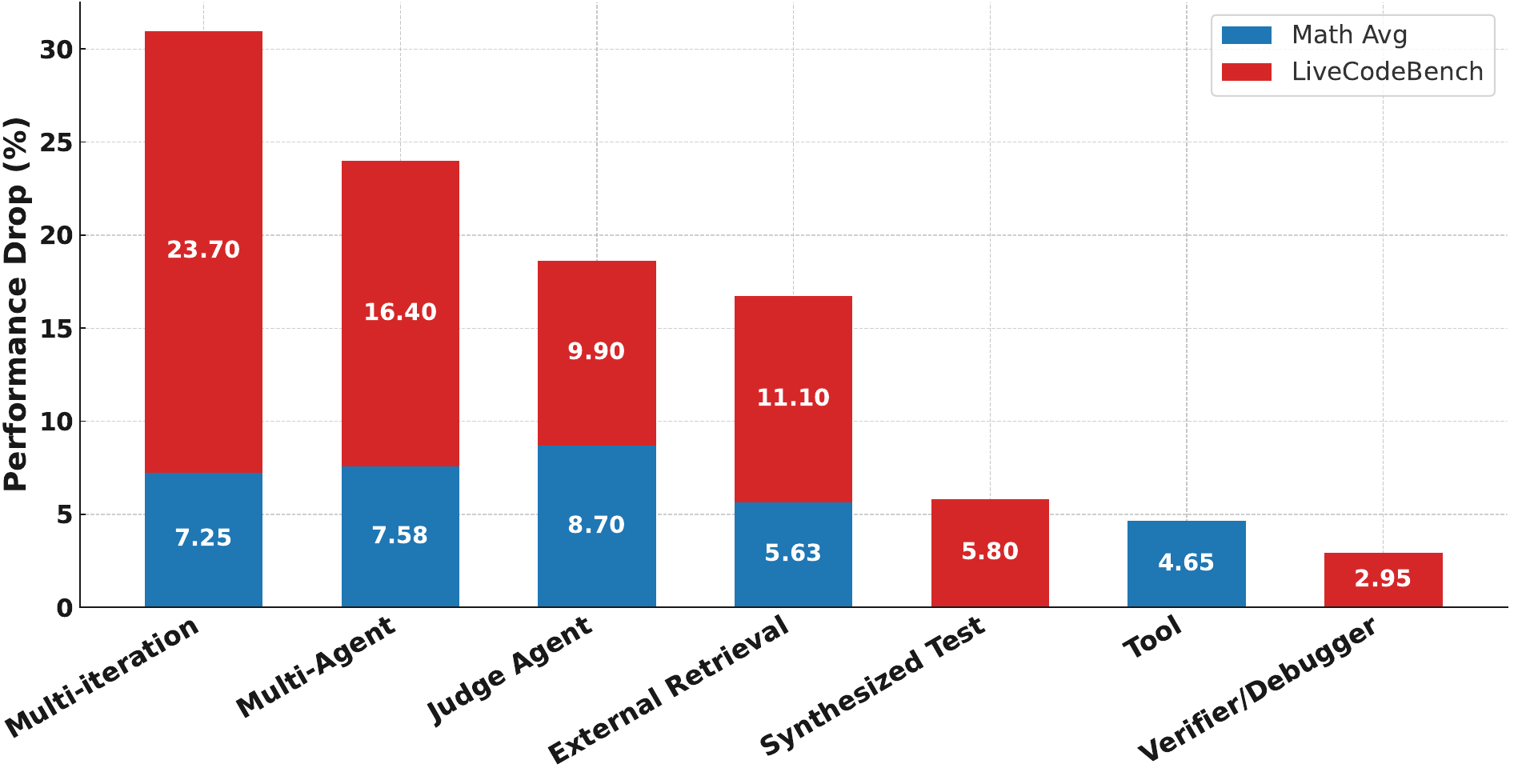}
    \vspace{-15pt}
    \caption{Performance drop when removing each component from \xolver. Bars show average drop on Math (bottom) and LiveCodeBench (top). Verifier is critical for math tasks and cannot be removed, while Tool (Python) and test cases apply only to math and coding respectively. }
    \vspace{-12pt}
    \label{fig:ablation-component-drop}
\end{wrapfigure}

\noindent{\bf Ablations: Quantifying Component Impact}
In Figure \ref{fig:ablation-component-drop}, we present an ablation study quantifying the contribution of individual components in \method{\xolver} to overall performance, measured by the average performance drop on math reasoning (Math Avg) and programming (LiveCodeBench) tasks. 
Each component plays a necessary role, with the most significant degradation observed when removing Multi-iteration and Multi-Agent followed by Judge Agent, highlighting their central importance in complex reasoning and code synthesis. 
In contrast, removing components like Verifier/Debugger and Tool leads to comparatively smaller drops, suggesting a more auxiliary role in the overall system. Likewise self-retrieval can also work in-place of external retrieval with some drop in accuracy.

\begin{wrapfigure}[]{r}{0.64\textwidth}
  \vspace{-15pt}
  \centering
  \includegraphics[width=\linewidth]{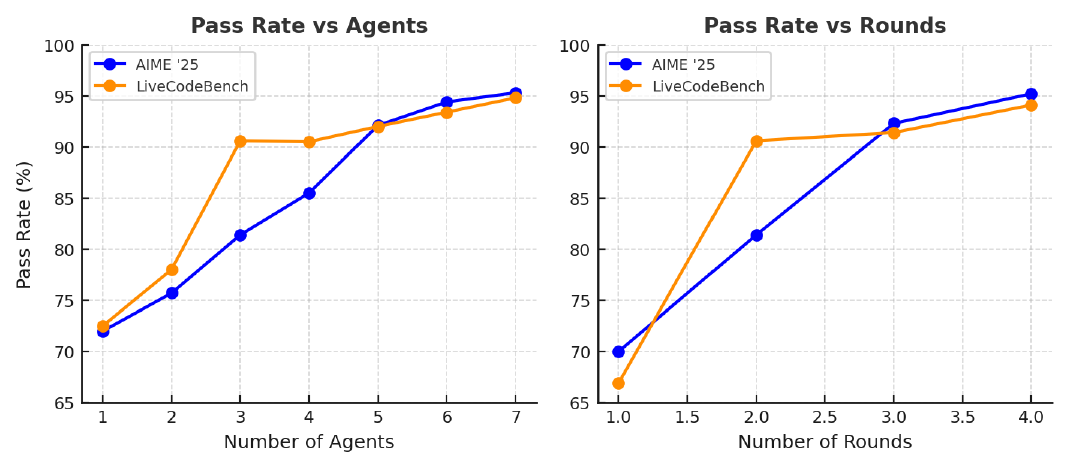}
  \vspace{-15pt}
  \caption{Impact of iterations and agents in \xolver on \textsc{AIME '25} (QWQ-32B) and \textsc{LiveCodeBench} (o3-mini-medium).} 
  \label{fig:appendix:iter_agent}
  \vspace{-5pt}
\end{wrapfigure}
\noindent\textbf{Impact of Agent Count and Iterations, and Emerging Benefits of Collaboration}  
We analyze the effect of varying the number of agents and reasoning iterations on \xolver's performance. In a controlled setup, we fix one variable (e.g., 3 agents or 2 iterations) and incrementally increase the other. As shown in Figure~\ref{fig:appendix:iter_agent}, performance improves consistently on both \textsc{AIME '25} and \textsc{LiveCodeBench} with more agents or iterations, highlighting the advantage of collaborative and iterative problem solving.

To probe deeper, we conduct a budget-controlled experiment on the \textsc{AIME '25} dataset, where the total reasoning budget (i.e., number of agents $\times$ number of iterations) is fixed. While iterative reasoning remains a crucial factor for \xolver's performance, we find that increasing the number of agents—particularly beyond a minimum of three—yields additional, emergent improvements, leading to over a 4\% performance gain. This suggests that agent diversity and parallelism complement iterative depth, together producing stronger collaborative problem-solving benefits than either alone.

\begin{wrapfigure}{hr}{0.55\textwidth}
    \vspace{-15pt}
    \centering
    \includegraphics[width=\linewidth]{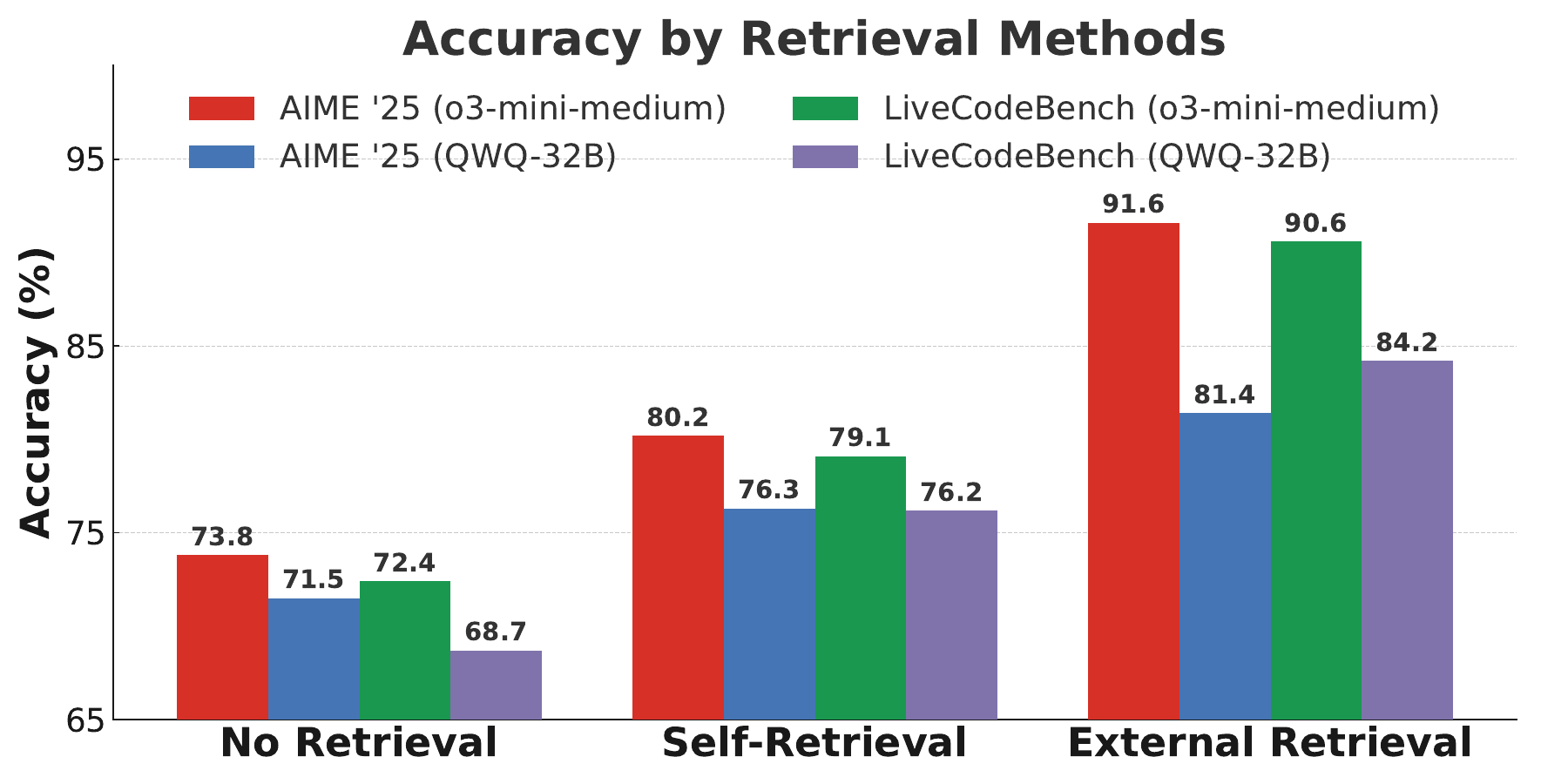}
    \vspace{-10pt}
    \caption{Impact of different retrievals  in \xolver.}
    \vspace{-10pt}
    \label{fig:appendix-rweteiavl-methods}
\end{wrapfigure}
\noindent\textbf{Effect of Retrieval Strategies on \xolver Performance.}  
We evaluate the impact of different retrieval strategies on \xolver by comparing three settings: (1) \textit{External Retrieval}, where the model retrieves the top-$k$ (e.g., $k=5$) most similar problems and their solutions from an external corpus using a BM25 retriever; (2) \textit{Self-Retrieval}, where the model recalls the top-$k$ most similar problems and solutions from its own internal memory; and (3) \textit{No Retrieval}, where neither external nor self-retrieval is used. 
As shown in Figure \ref{fig:appendix-rweteiavl-methods}, performance on both \textsc{AIME '25} and \textsc{LiveCodeBench} follows the trend: \textit{External Retrieval} $>$ \textit{Self-Retrieval} $>$ \textit{No Retrieval}, indicating that external retrieval significantly enhances \xolver's performance. 
We note that for code tasks, although the external retrieval corpus contains solutions written in C++—a different language from the target Python—external retrieval still provides a substantial performance boost. Nonetheless, while self-retrieval results in a notable performance drop compared to external retrieval, it still outperforms the no-retrieval baseline with notable margins, serving as a viable alternative when external resources are unavailable.

\begin{wrapfigure}[]{r}{0.68\textwidth}
  \vspace{-15pt}
\centering
  \includegraphics[width=\linewidth]{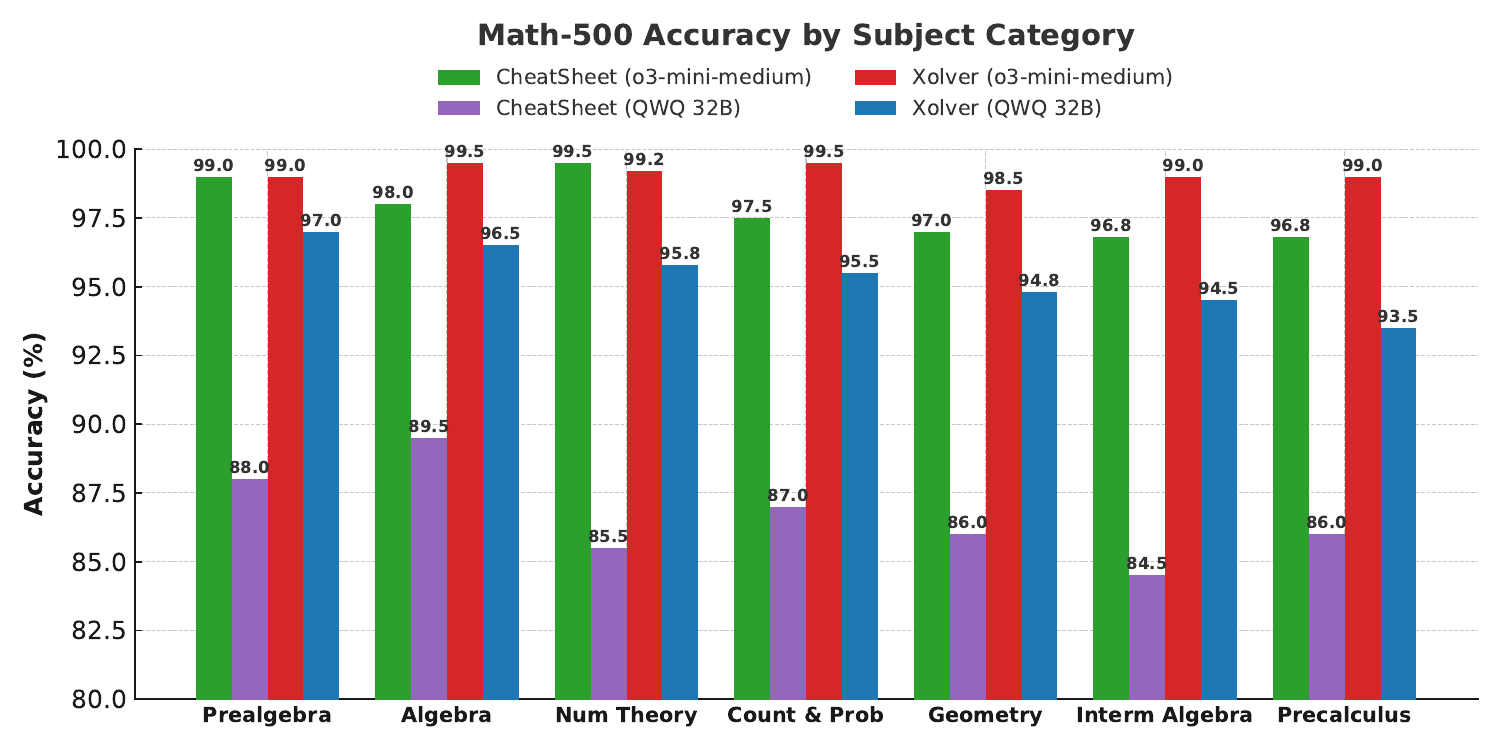}
  \vspace{-15pt}
  \caption{Fine-grained performance comparison in MATH-500.} 
  \label{fig:appendix-fine-grained}
  \vspace{-10pt}
\end{wrapfigure}
\begin{wrapfigure}[]{r}{0.4\textwidth}
  \vspace{-25pt}
 \hspace{1pt}
  \includegraphics[width=\linewidth]{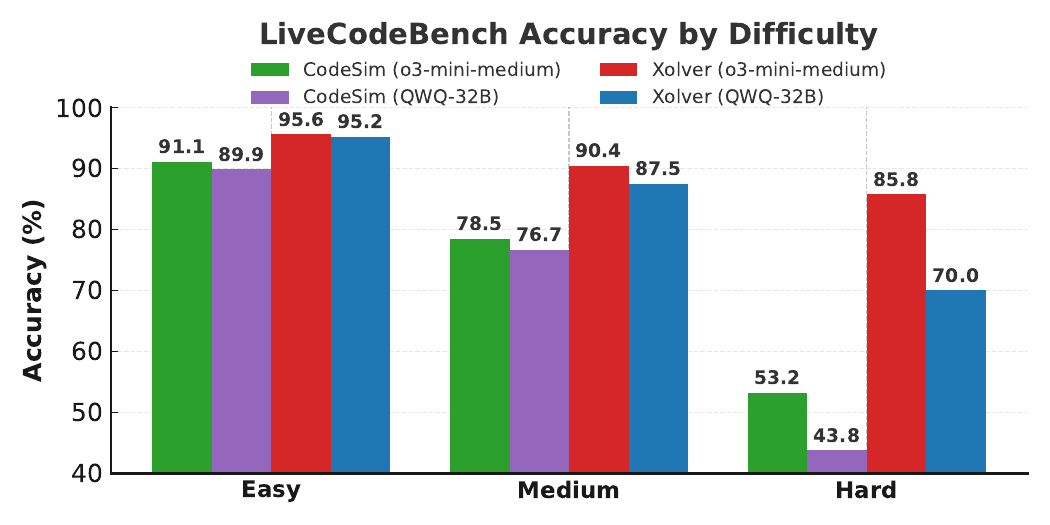}
  \vspace{-15pt}
 \caption{Performance comparison per difficulty levels in LiveCodeBench}
    \label{fig:appendix-fine-grained-code}
  \vspace{-10pt}
\end{wrapfigure}

\noindent{\bf Fine-grained Performance Analysis} 
We perform a fine-grained analysis of \method{\xolver}’s performance across both \textsc{Math-500} and \textsc{LiveCodeBench}, as shown in Figure~\ref{fig:appendix-fine-grained} and Figure~\ref{fig:appendix-fine-grained-code}.On \textsc{Math-500}, \xolver (both {o3-mini-medium} and {QWQ-32B}) consistently outperforms \textsc{CheatSheet} across nearly all seven subject categories, despite the latter relying on costly per-problem memory updates. The only exception is in \textit{Number Theory}, where \texttt{o3-mini-medium} scores 99.2 compared to \textsc{CheatSheet}'s 99.5. As for QWQ-32B,
\xolver achieves substantial accuracy gains over CheatSheet across all categories, with improvements of +9.0\% in Prealgebra, +8.5\% in Algebra, +11.0\% in Number Theory, +8.5\% in Counting and Probability, +8.8\% in Geometry, +10.0\% in Intermediate Algebra, and +7.5\% in Precalculus. These consistent gains highlight \method{\xolver}’s strong performance across both symbolic and numerical reasoning.


On LiveCodeBench, \xolver demonstrates even more pronounced gains. The o3-mini-medium variant achieves 95.6\%, 90.4\%, and 85.8\% accuracy on Easy, Medium, and Hard problems respectively, significantly outperforming CodeSim by +4.5\%, +11.9\%, and a striking +32.3\% margin on hard examples. Even with a weaker QWQ-32B backbone, \xolver (95.2\%, 87.5\%, 70.0\%) surpasses all baselines and achieves similar gains. In contrast to CheatSheet and CodeSim, \xolver leverages multi-agent collaborations and holistic experience learning. These consistent and backbone-agnostic gains across different reasoning tasks underscore \method{\xolver}’s robustness and position it as a breakthrough in retrieval and tool-augmented, multi-agent and evolving reasoning systems.

\noindent{\bf Can a Self-Judge Replace a Judge Agent?} We analyze the effect of different judging mechanisms on \xolver's performance by comparing two setups: (1) \emph{self-judging}, where each dynamic agent evaluates its own response through self-reflection without altering its role, and (2) \emph{external judging}, where a separate judge agent is used to assess the responses. We find that self-judging agents tend to be biased in favor of their own outputs, occasionally validating incorrect solutions. This self-bias leads to a noticeable drop in overall performance---specifically, a {9.9\% decrease in coding tasks} and a {3.88\% decrease in math tasks}, on average.

\begin{wrapfigure}{r}{0.7\textwidth}
    \vspace{-15pt}
    \centering
    \includegraphics[width=\linewidth]{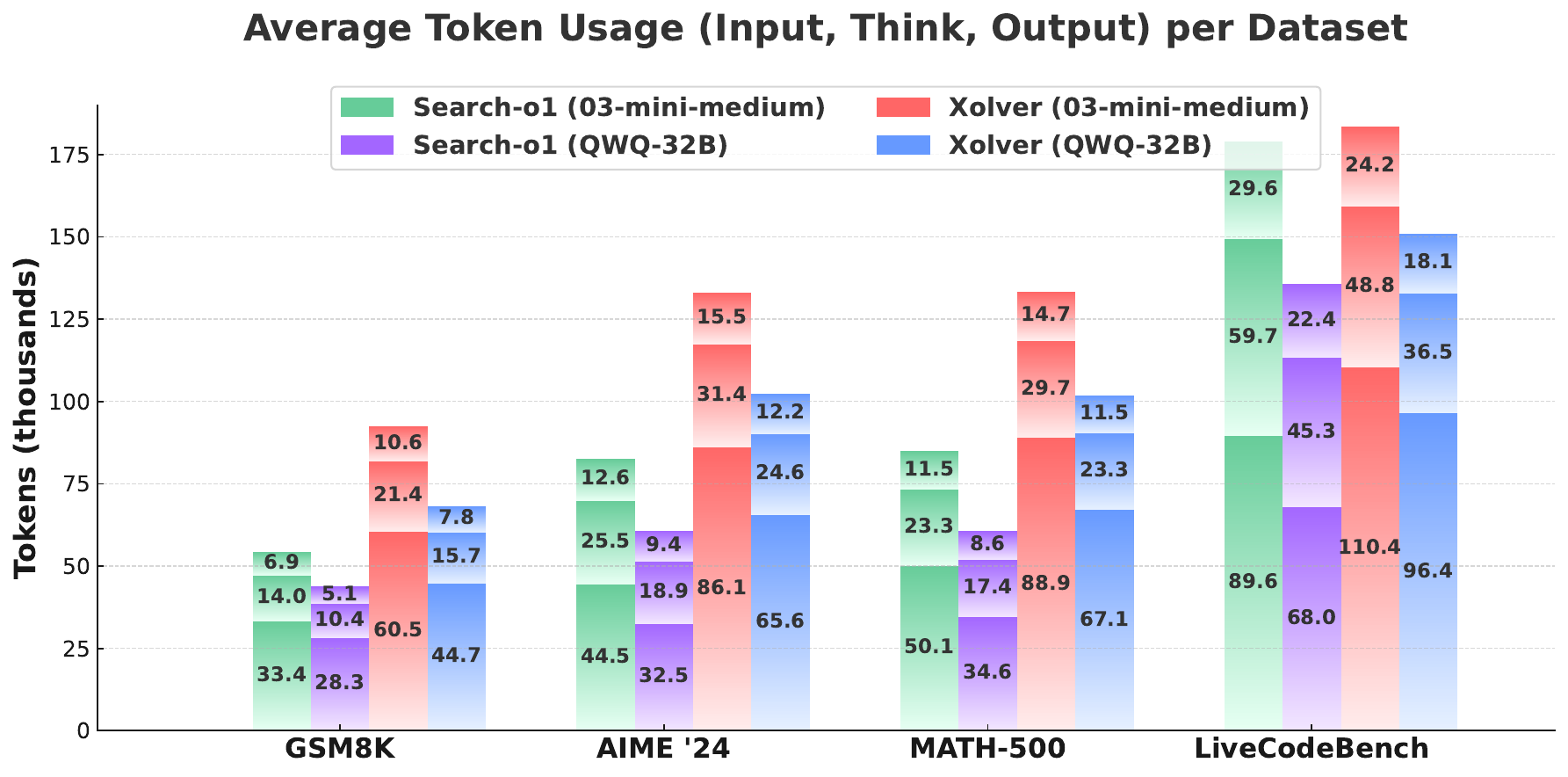}
    \vspace{-18pt}
    \caption{Avg numbers of token usage across datasets in \xolver (+).}
    \vspace{-15pt}
    \label{fig:appendix-token}
\end{wrapfigure}

\noindent {\bf Cost Analysis and How Long Do \xolvercolor Agents Think?}
We perform a detailed analysis of token usage in Figure~\ref{fig:appendix-token}, reporting input, reasoning, and output statistics for \xolver~(QWQ-32B) across all datasets. Our LLM token usage has computational complexity of $O(m \mathcal{I})$, where $m$ is the number of agents and $\mathcal{I}$ is the number of reasoning iterations. However, the runtime complexity remains $O(\mathcal{I})$ since the dynamic agents operate in parallel. This is significantly more efficient than the self-consistency \citep{self_consistency}, which typically require 32–64 generations per example, as well as the baseline CheatSheet framework, which incurs a memory update complexity of $O(n^2)$—quadratic in the test dataset size—due to usefulness estimation over all previous examples after solving each new example. As a multi-agent system, \xolver allocates a majority of its tokens to context sharing and inter-agent communication, while approximately 25\% are spent on actual reasoning steps.

Nonetheless in Figure~\ref{fig:appendix-token}, we also compare the total token usage of \xolver with a single agent reasoning framework Search-o1 using \texttt{tiktoken} for o3-mini-medium and \texttt{AutoTokenizer} for QWQ-32B for token count. As expected, \xolver incurs higher token costs—approximately $1.5\times$ that of Search-o1—due to its collaborative and iterative multi-agent reasoning. However, this moderate increase represents a highly efficient trade-off given the substantial performance improvements observed. As shown in Figure \ref{fig:appendix-fine-grained} and Figure \ref{fig:appendix-fine-grained-code}, \xolver achieves remarkable gains across both domains, including a +32.3\% absolute improvement on hard coding problems with o3-mini-medium and 9.05\% accuracy boosts across all Math-500 categories with QWQ-32B. These findings demonstrate that \xolver’s slightly higher reasoning cost is well-justified by its superior, generalist performance across diverse problem-solving scenarios.

\noindent\textbf{Does Data Shuffling Affect \xolver (+) Performance?}  
\xolver (+) updates its external memory incrementally after solving each new problem. To examine whether the order of test instances impacts performance, we conduct an ablation study by randomly shuffling the sequence of problems in each task. This helps determine if there is any dependency on the data order. Results in Appendix~\ref{sec:appendix:shufflexx} show that \xolver exhibits minimal performance variation across different shuffles, with a standard deviation of approximately $1$ within only 5 runs, indicating that its performance is largely stable regardless of data ordering.

\noindent {\bf Qualitative Examples} In Appendix~\ref{sec:appendix:qual_exmp}, we present qualitative examples along with all the prompts of full-cycle \xolver on both math and code reasoning tasks. These examples illustrate how \xolver initiates reasoning from external or self-retrieved exemplars, engages in multi-agent collaboration, and incrementally accumulates experiences through inter-agent propagation and refinement. The full interaction trace highlights \xolver's ability to iteratively decompose, solve, and adapt solutions across reasoning steps, showcasing its capacity for dynamic knowledge construction and generalizable problem solving.

\begin{wrapfigure}[]{r}{0.65\textwidth}
  \vspace{-15pt}
\centering
  \includegraphics[width=\linewidth]{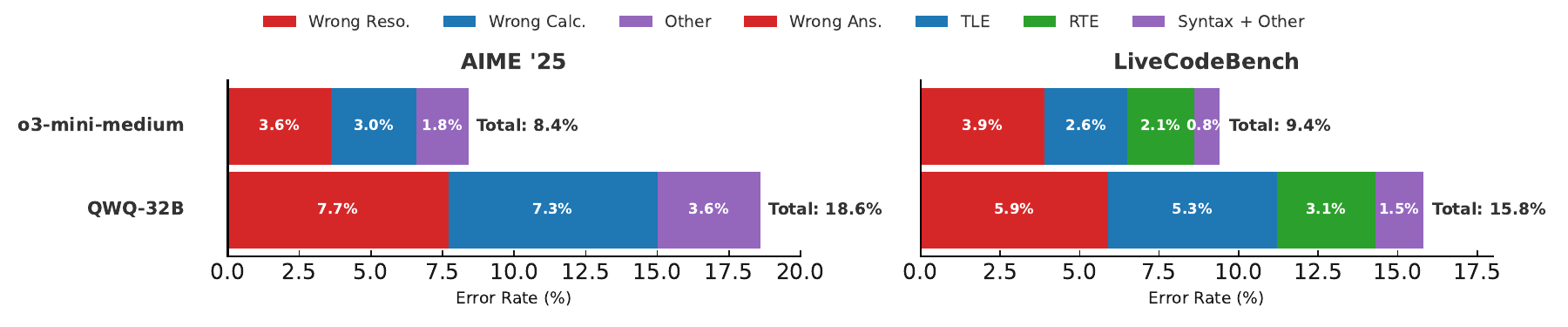}
  \caption{\xolver Math and Code error distribution. } 
  \label{fig:xolver-error-analysis}
  \vspace{-15pt}
\end{wrapfigure}
\noindent {\bf More Error Analysis in Math and Code} 
In Figure~\ref{fig:xolver-error-analysis}, we present an error analysis across both math and code tasks that goes beyond simple accuracy or pass@1 metrics. While \xolver significantly improves reasoning and generation capabilities in both domains, both (o3-mini-medium and QWQ-32B) backbone LLMs can still produce solutions that are syntactically correct yet semantically flawed, resulting in failed executions due to incorrect reasoning, incomplete logic, unoptimized implementations, or misaligned tool usage. In code tasks, failure modes include incorrect final code, time limit exceeded (TLE), runtime errors (RTE), and syntax issues. In math tasks, remaining errors are primarily due to flawed logical derivations or faulty intermediate calculations. Although Python-based tools are available, such calculation errors often occur when agents choose not to invoke these tools—highlighting that tool usage remains decoupled from the model’s core reasoning process (see Appendix~\ref{sec:appednix:prompts} for our prompt design). These findings provide insights for future improvements by exposing the variety of failure modes across domains, and further emphasize the importance of robust self-verification and refinement mechanisms, as employed by \xolver.

\noindent{\bf Dynamics of Reasoning Patterns in \xolver Traces}
\label{sec:appendix:reasoning}
To understand how \xolver adapts its reasoning process to perform complex reasoning, we analyze the dynamics of reasoning pattern frequencies across difficulty levels in LiveCodeBench, as shown in Table~\ref{tab:reasoning_trace}. Detailed description of how we collected the
reasoning patterns is provided in the Appendix \ref{sec:appendix:reasoning}. Our analysis reveals that \xolver dynamically increases \emph{self-evaluation} and \emph{exploratory strategies} (e.g., trying new approaches) as problem difficulty grows. Correct solutions demonstrate a declining need for problem rephrasing and subgoal decomposition, indicating more direct and confident reasoning. In contrast, incorrect solutions show increased subgoal setup and rephrasing attempts—suggesting that the system recognizes failure and attempts recovery through restructuring. Compared to OpenCodeReasoning, which shows stagnation or regression in key patterns (e.g., self-evaluation), \xolver exhibits robust and adaptive reasoning behavior, supported by multi-agent collaboration and judge feedback. This behavior highlights the generality and flexibility of \xolver’s reasoning model.

\begin{table}[t]
\centering
\adjustbox{max width=\textwidth}{
\begin{tabular}{lcccc}
\hline
\textbf{Reasoning Pattern} 
& \multicolumn{2}{c}{\textbf{Correct Solutions}} 
& \multicolumn{2}{c}{\textbf{Incorrect Solutions}} \\
& Easy $\rightarrow$ Medium & Medium $\rightarrow$ High 
& Easy $\rightarrow$ Medium & Medium $\rightarrow$ High \\
\hline
Self-Evaluation ($\uparrow$)           
& \highlightcell{\underline{\textcolor{OliveGreen}{0.35 $\rightarrow$ 0.38}}} & \highlightcell{\underline{\textcolor{OliveGreen}{0.38 $\rightarrow$ 0.40}}} 
& \underline{0.35 $\rightarrow$ 0.37} & \highlightcell{\underline{\textcolor{OliveGreen}{0.32 $\rightarrow$ 0.35}}} \\
New Approach ($\uparrow$)
& {\textcolor{OliveGreen}{0.18 $\rightarrow$ 0.21}} & {\textcolor{OliveGreen}{0.21 $\rightarrow$ 0.24}} 
& {\textcolor{OliveGreen}{0.17 $\rightarrow$ 0.24}} & {\textcolor{OliveGreen}{0.24 $\rightarrow$ 0.26}} \\
Problem Rephrasing ($\downarrow\uparrow$)         
& \textcolor{red}{0.20 $\rightarrow$ 0.17} & 0.18 $\rightarrow$ 0.18
& 0.23 $\rightarrow$ 0.24 & \textcolor{OliveGreen}{0.24 $\rightarrow$ 0.25} \\
Subgoal Setup ($\downarrow\uparrow$)
& \textcolor{red}{0.14 $\rightarrow$ 0.13} & \textcolor{red}{0.13 $\rightarrow$ 0.11} 
& \highlightcell{\underline{\textcolor{OliveGreen}{0.11 $\rightarrow$ 0.12}}} & \underline{0.11 $\rightarrow$ 0.11} \\
\hline
\end{tabular}
}
\vspace{2pt}
\caption{
Changes in major reasoning pattern frequencies as problem difficulty increases in LiveCodeBench, comparing correct vs. incorrect solutions. \textcolor{OliveGreen}{Green} and \textcolor{red}{red} indicate statistically significant increases or decreases ($p < 0.05$). Underlined cells highlight patterns where \xolver improves over OpenCodeReasoning, which otherwise shows a declining trend. Direction arrows denote: 
$\uparrow$ = increase, 
$\downarrow$ = decrease, 
$\downarrow\uparrow$ = mixed trend (decrease in correct, increase in incorrect). 
\xolver increases use of self-evaluation and new approaches with task difficulty, and demonstrates targeted subgoal setup and problem rephrasing when solutions fail—reflecting its adaptive, collaborative reasoning.
}
\label{tab:reasoning_trace}
\vspace{-20pt}
\end{table}
\section{Case-Study: How \xolver Enhances Reasoning}
\begin{wrapfigure}[]{r}{0.65\textwidth}
  \vspace{-15pt}
\centering
  \includegraphics[width=\linewidth]{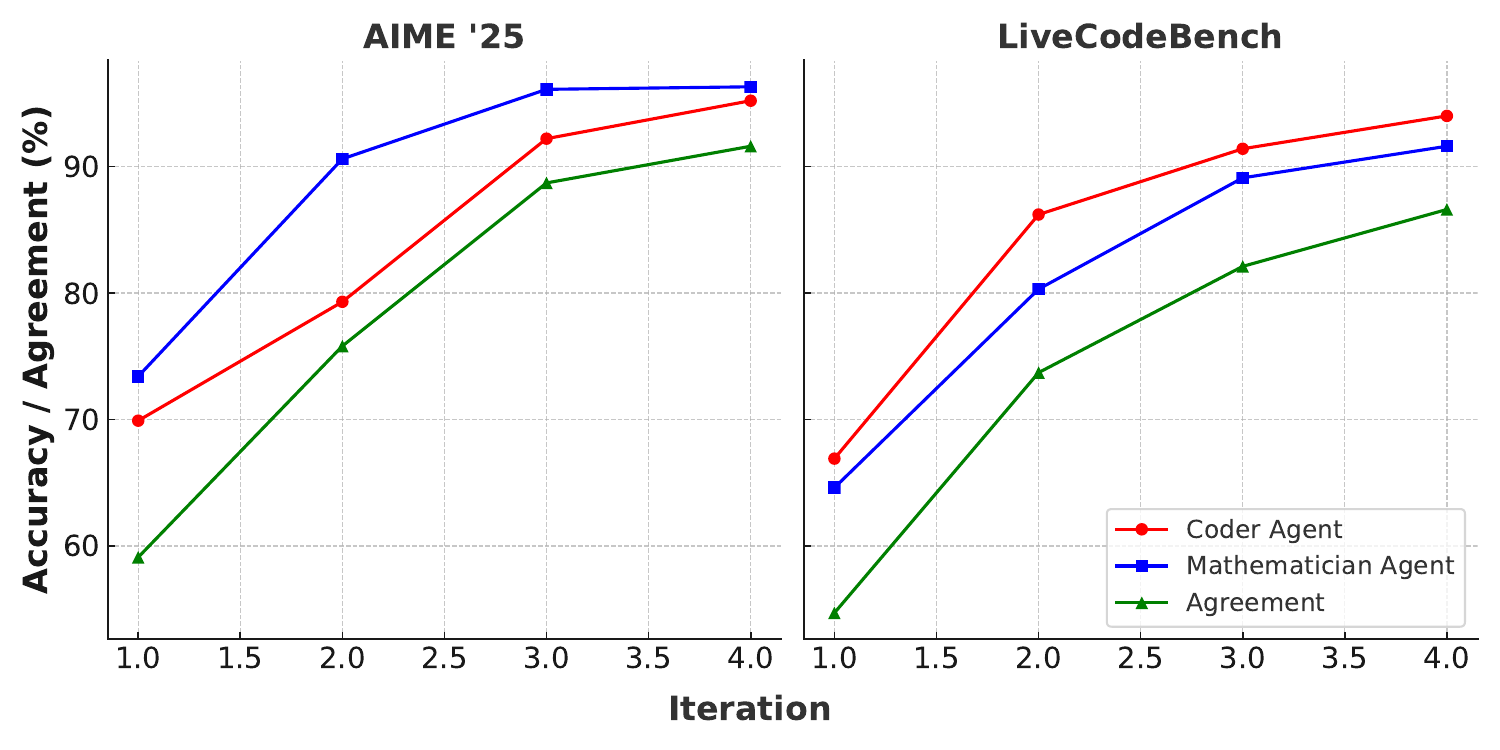}
  \vspace{-20pt}
  \caption{Agents Accuracy and Agreement over iterations.} 
  \label{fig:appendix-aggrement}
  \vspace{-10pt}
\end{wrapfigure}
To further understand the reasoning and problem-solving strategies behind our multi-agent, iterative framework \xolver, we conduct an in-depth analysis combining qualitative runtime inspection with controlled experiments. We begin by manually studying \method{\xolver}’s agent interaction traces on AIME '25 and LiveCodeBench. These case studies reveal that at each iteration, dynamic agents attempt to improve upon earlier failures by leveraging Judge agent feedback and by aligning with top-ranked outputs stored in the shared memory $\mathcal{D}_S$. This process results in progressively refined outputs, increased agent alignment, and eventual convergence toward correct solutions.

To verify this behavior systematically, we conduct a controlled experiment across both math and code tasks. We instantiate two dynamic agents with complementary strengths: a Coder agent and a Mathematician agent, each proficient in one domain but suboptimal in the other. We then measure their performance and agreement across iterations—defined as the percentage of problems in which both agents independently produce the same correct answer (for math) or code that passes the same test cases (for code). As shown in Figure~\ref{fig:appendix-aggrement}, both agents demonstrate consistent accuracy improvements over time, accompanied by a rising agreement rate. This not only illustrates mutual influence and learning-by-alignment but also validates the emergence of collaborative synergy.

Crucially, we observe that the presence of the Judge agent plays a vital role in this convergence process. When the Judge agent is removed—as shown in our first ablation—performance degrades significantly. These findings collectively affirm that \method{\xolver}’s iterative memory-sharing, feedback-driven refinement, and role-specialized agents contribute to its strong reasoning performance across domains, making it a compelling framework for general-purpose, self-improving problem solving.

\section{Related Work}
\label{sec:related_work}

\textbf{Memory-Augmented and Retrieval-Augmented LLMs.} Memory-augmented language models have evolved from static retrieval systems like RAG~\citep{lewis2020retrieval} and REALM~\citep{guu2020realm} to dynamic approaches such as Reflexion~\citep{shinn2024reflexion}, MemGPT~\citep{Packer2023MemGPTTL}, and Scratchpads~\citep{nye2021show}. However, these systems operate on isolated tasks, lack cross-problem experience accumulation, and employ single-agent architectures. \method{\xolver} addresses these limitations through a novel dual-memory architecture combining episodic long-term memory with dynamic intermediate memory, enabling specialized agents to collectively build and refine experiential knowledge. While prior work has explored cross-trial information sharing~\citep{Zhao2023ExpeLLA,shinn2024reflexion} and multi-source memory integration~\citep{Yao2023RetroformerRL}, these approaches remain confined to single-agent settings. Our framework creates a persistent knowledge base through multi-agent collaboration~\citep{Feng2025WhenOL}, allowing agents to accumulate expertise from solved problems and leverage collective experience for future tasks.

\textbf{Multi-Agent Problem Solving.} Multi-agent LLM systems address the limitations of single models by leveraging collaborative approaches for improved reliability and task specialization~\citep{guo2024large, Feng2025WhenOL}. From early frameworks like CAMEL~\citep{li2023camel} with fixed role assignments, the field progressed to dynamic role adjustment in AgentVerse~\citep{chen2023agentverse} and code execution in AutoGen~\citep{wu2023autogen}. Recent advances include layered agent networks in DyLAN~\citep{liu2023dynamic}, multi-agent code generation and problem solving \citep{islam-etal-2024-mapcoder, islam-etal-2025-codesim} and multi-agent debate frameworks~\citep{du2023improving, rahman2025ai, subramaniam2025multiagent}. While these systems demonstrate effective collaboration, they operate on isolated problems without cross-task experience accumulation. \method{\xolver} introduces dual-memory architecture, holistic experience integration, judge-mediated selection, and continuous episodic corpus expansion—transforming single-problem solvers into experience-aware agents.

\textbf{LLM Reasoning Enhancement Techniques.} Various techniques have emerged to enhance LLM reasoning capabilities beyond standard prompting. Chain-of-Thought~\citep{wei2022chain} introduced step-by-step reasoning, Self-Consistency~\citep{wangself} explores multiple reasoning paths with majority voting, and Tree of Thoughts~\citep{yao2023tree} enables exploration of reasoning branches—yet all remain limited to single-pass generation. Self-reflective approaches like Reflexion~\citep{shinn2024reflexion} enable iterative improvement but operate within single tasks, while retrieval-enhanced methods like CheatSheet~\citep{suzgun2025dynamic} and Search-o1~\citep{li2025search} remain confined to single-agent architectures. These approaches share fundamental limitations: no cross-problem learning, no persistent memory, and no multi-agent collaboration. \method{\xolver} unifies these enhancements within a multi-agent framework where agents collaboratively refine solutions through judge-mediated iterations and leverage dual memory systems for cross-problem learning.

\textbf{Tool-Augmented Reasoning.} Tool integration extends LLM capabilities beyond language processing. Early systems like WebGPT~\citep{nakano2021webgpt} introduced single-tool integration, while PAL~\citep{gao2023pal} enabled code execution for mathematical reasoning. Multi-tool frameworks evolved with ReAct~\citep{yao2023react} interleaving reasoning with actions, Chameleon~\citep{lu2024chameleon} composing multiple tools, and OctoTools~\citep{lu2025octotools} standardizing tool planning—yet all remain limited to single-agent execution without iterative refinement or cross-problem learning. \method{\xolver} transforms tool use into a collaborative, memory-enriched ecosystem where agents collectively execute tools, share outcomes, and accumulate successful strategies across problems—creating an adaptive framework that evolves with experience.

\section{Conclusion}
\label{sec:conclusion}
We propose \xolver, an open-source, multi-agent inference framework for complex reasoning tasks that enables holistic experience learning. \xolver integrates (1) episodic retrieval from external or self-parametric memory, (2) an evolving intermediate shared memory that accumulates and reuses high-quality reasoning traces, (3) tool invocation for complex computations, (4) collaborative multi-agent reasoning, (5) self-evaluation and iterative refinement, (6) verification or external debugging, and (7) propagation of learned strategies across problems. These components collectively support adaptive, experience-informed problem solving.
Despite its strong performance, \xolver faces limitations in computational efficiency, with substantially higher token consumption than traditional approaches, and remains dependent on the quality of backbone LLMs. Future work aims to optimize agent interactions to reduce resource requirements, enhance robustness to variations in model quality, improve retrieval filtering \citep{filco, sadat-etal-2023-delucionqa, parvez-chang-2021-evaluating}, develop better RAG strategies \citep{islam-etal-2024-open, parvez-etal-2023-retrieval, lekssays2025techniqueragretrievalaugmentedgeneration}, and extend the framework to more diverse reasoning domains beyond mathematics and programming. In addition, we plan to integrate advanced external verifiers of reasoning~\citep{mahan2024generative} to further enforce validity through structured guardrails. By addressing these challenges, we aim to further advance the development of experience-aware reasoning systems that can approach the adaptability and integrated knowledge use of human experts.



\clearpage
\bibliography{Reference, custom}


\bibliographystyle{plainnat}


\clearpage

\appendix
\clearpage

\startcontents[sections]
\printcontents[sections]{l}{1}{\setcounter{tocdepth}{3}}

\clearpage

\section{Lists of Prompts}
\label{sec:appednix:prompts}
This section provides the list of prompts for planning, dynamic, judge, verifier and reasoning segmentation we have used in the experimental period. These are crucial to ensure the reproducibility \citep{laskar-etal-2024-systematic} of the framework \method{\xolver}. 
\subsection{Planner Agent}

\begin{tcolorbox}[title=Prompt for \textsc{Planner Agent}, colback=blue!5!white, colframe=blue!25!black, fonttitle=\footnotesize, boxrule=0.2mm, sharp corners]
\label{fig:prompt_response_break_down_example}
\small
You are a planner to solve a \{coding/math\} problem. Here is the problem for which you have to plan:\\
\texttt{\{problem\_dict[query[`problem\_id']][`description']\}}\\

First draft required strictly greater than \texttt{\{m\}} specialized roles to solve the problem collaboratively with reasoning behind your draft of each role.  \\

Then select the highly influential \texttt{\{m\}} roles by re-checking the reasoning behind your selection and assign them to each agent to solve the problem.
\end{tcolorbox}

\subsection{Dynamic Agent}

\begin{tcolorbox}[title=Prompt for \textsc{Dynamic Agent}, colback=blue!5!white, colframe=blue!25!black, fonttitle=\footnotesize, boxrule=0.2mm, sharp corners]
\label{}
\small
You are a \texttt{\{role\}}. Your task is to solve a \{coding/math\} problem. Here is the problem that you have to solve:\\
\texttt{problem\_dict[query[`problem\_id']][`description']}\\
------------------------------------------------------------------------------------------------------------------  \\
\textbf{If external retrieval:}
You were also given a couple of similar problems to the problem above along with their reasoning and solutions to aid you in solving the problem at hand. Here are the similar problems you were given:\\
\texttt{retrieved\_dict[query[`problem\_id']][`retrieval\_text']}\\
\newline
\textbf{If self-retrieval:}
Further, recall a relevant and distinct problem (different from the problem mentioned above) along with its reasoning and solution.\\
------------------------------------------------------------------------------------------------------------------  \\
And here was your original response:\\
\texttt{query[[`role']][`original\_thought',`original\_response']}\\
------------------------------------------------------------------------------------------------------------------  \\
\textbf{If iteration $i\geq1$ (i.e., $\mathcal{D}_S$ is not empty):} \\
Also here is the leading responses with execution results from the response store:\\
\texttt{response\_dict[`role',`thought',`response',`score']}\\
------------------------------------------------------------------------------------------------------------------  \\
\textbf{If coding task:}\\
Think carefully about where you went wrong, relating with responses in the response store. Then, try to fix the solution producing a thought later reply with a \{Python\} solution to be executed and judged again. \\

Make sure to wrap your code in \texttt{\textasciigrave\textasciigrave\textasciigrave python \textasciigrave\textasciigrave\textasciigrave} block and \texttt{} Markdown delimiters, and include exactly one block of code with the entire solution (in the final code step).\\
------------------------------------------------------------------------------------------------------------------  \\
\textbf{If math task:}\\
Think carefully about where you went wrong, relating with responses in the response store. Then, try to fix the solution producing a thought later reply with a solution to be executed and judged again. You can integrate a \{Python\} tool to execute the calculations after replying your solution if required.\\

Make sure to wrap your final answer in \texttt{\textbackslash boxed\{\}} block with the entire solution (in the final answer step).\\
\end{tcolorbox}

\subsection{Judge Agent}

\begin{tcolorbox}[title=Prompt for \textsc{Judge Agent}, colback=blue!5!white, colframe=blue!25!black, fonttitle=\footnotesize, boxrule=0.2mm, sharp corners]
\label{}
\small
You are a judge. Your task is to judge the candidate solution of a \{coding/math\} problem. Here is the problem for which the candidate solution you have to judge:\\
\texttt{problem\_dict[query[`problem\_id']][`description']}\\
------------------------------------------------------------------------------------------------------------------  \\
\textbf{If coding task:}\\
And here is the candidate response along with test cases against which to judge:\\
\texttt{query[[`candidate\_role']][`candidate\_thought',`candidate\_response',`test\_case']}\\

Please produce a score (based on the number of test cases passed) with reasoning behind your judgement of the candidate solution to the problem.\\
------------------------------------------------------------------------------------------------------------------  \\
\textbf{If math task:}\\
And here is the candidate response which to judge:\\
\texttt{query[[`candidate\_role']][`candidate\_thought',`candidate\_response'}\\

Please produce a score (if the response is correct, it should be 1 otherwise should be 0) with reasoning behind your judgement of the candidate solution to the problem.
\end{tcolorbox}

\subsection{Verifier Agent}

\begin{tcolorbox}[title=Prompt for \textsc{Verifier Agent}, colback=blue!5!white, colframe=blue!25!black, fonttitle=\footnotesize, boxrule=0.2mm, sharp corners]
\label{}
\small
Your are an answer extractor. Your task is to extract answer from the response to a \{coding/math\} problem. Here is the response for which the answer you have to extract:\\
\texttt{response\_dict[query\{`role'\}]\{`thought', `response', `score'\}}\\
------------------------------------------------------------------------------------------------------------------  \\
\textbf{If coding task:}\\
Please extract the answer from inside \texttt{\textasciigrave\textasciigrave\textasciigrave python \textasciigrave\textasciigrave\textasciigrave} block from the response.\\
------------------------------------------------------------------------------------------------------------------  \\
\textbf{If math task:}\\
Please extract the answer from inside \texttt{\textbackslash boxed\{\}} block from the response.\\
\end{tcolorbox}

\subsection{Reasoning Segmentation}
\label{sec:appendix:reaso_seg}
\begin{tcolorbox}[title=Prompt for \textsc{Initial Reasoning Segmentation}, colback=blue!5!white, colframe=blue!25!black, fonttitle=\footnotesize, boxrule=0.2mm, sharp corners]
\label{}
Below is a chain of thought for solving a question. Figure out what are the different reasoning patterns that are used like problem rephrasing, new approach, self-evaluation, verification,
backtracking, subgoal setup, solving a simpler problem first, and more. Then your task is to segment the entire chain of thought into different reasoning patterns. Rewrite the chain of thought in the following format:\\
\begin{verbatim}
<pattern> pattern name </pattern>
<content> the entire text that corresponds to the pattern </content>
\end{verbatim}
Thought: \{thoughts\}
\end{tcolorbox}

\begin{tcolorbox}[title=Prompt for \textsc{Final Reasoning Segmentation}, colback=blue!5!white, colframe=blue!25!black, fonttitle=\footnotesize, boxrule=0.2mm, sharp corners]
\label{}
Below is a thought for solving a question. For the segment between the
\texttt{<unannotated>} and \texttt{</unannotated>} tags, figure out what is reasoning pattern used in
that segment like problem rephrasing, new approach, self-evaluation, verification,
backtracking, subgoal setup, solving a simpler problem first, or something else.
Then your task is to identify the reasoning pattern used in the unannotated segment.
Rewrite the unannotated segment in the following format:\\

\begin{verbatim}
<content> The text within the unannotated segment that corresponds to
the pattern. </content>
<reasoning> Reasoning for what the pattern should be for the content 
</reasoning>
<pattern> *single* pattern name </pattern>
\end{verbatim}

Thought: \{thoughts\}
\end{tcolorbox}

\section{Qualitative Examples}
\label{sec:appendix:qual_exmp}
The subsequent section presents some qualitative examples--agentic interaction breakdown on how \xolver can approach and solve a problem of math and coding tasks when it encounters.

\begin{tcolorbox}[breakable, title=Example for \textsc{Correct Math Task \colorbox{OliveGreen}{Green}: Correct Answer. \colorbox{Red}{Red}: Wrong Answer.}, colback=blue!5!white, colframe=blue!25!black, fonttitle=\footnotesize, boxrule=0.2mm, sharp corners]
\label{}
\small

\colorbox{pink}{Query:} Bob is in charge of doing laundry for a large hotel. Each room has two sheets, one comforter, twice as many pillow cases as sheets and twice as many towels as pillow cases. How many pieces of laundry are there in 80 rooms? Make sure to wrap your final answer in \texttt{\textbackslash boxed\{\}} block with the entire solution (in the final answer step).\\

\noindent\hfill\makebox[0.6\linewidth]{\textbf{Iteration 1}}\hfill\null
\medskip 

\colorbox{Red}{Mathematical Modeler:}
Each room contains 2 sheets, 1 comforter, 2 pillow cases (which is twice the number of sheets), and 4 towels (which is twice the number of pillow cases). Hence, the total number of laundry items in each room is:\\
\(2 + 1 + 2 + 4 = 9\) pieces of laundry.\\
When considering 80 rooms, the total number of pieces of laundry is:\\
\(80 \times 9 = 720\) pieces of laundry.\\
Thus, the final answer is: \boxed{720}.\\

\colorbox{Yellow}{Judge:} The candidate's solution contains an error in calculating the number of pillow cases and towels per room. Here's the breakdown:\\
Sheets: 2 (correct). Comforter: 1 (correct). Pillow Cases: The problem states ``twice as many pillow cases as sheets.'' Since there are 2 sheets, pillow cases should be \(2 \times 2 = 4\). The candidate incorrectly used 2 pillow cases (half the required amount). Towels: The problem states ``twice as many towels as pillow cases.'' With the correct pillow case count (4), towels should be \(2 \times 4 = 8\). The candidate used 4 towels (based on their incorrect pillow case count of 2).\\
The correct per-room total is:\\
\(2 \text{ (sheets)} + 1 \text{ (comforter)} + 4 \text{ (pillow cases)} + 8 \text{ (towels)} = 15\)\\
items. The total for 80 rooms is:\\
\(80 \times 15 = 1200\).\\

The candidate's answer is incorrect due to miscalculating pillow cases and towels.\\

Score: 0.\\

\colorbox{OliveGreen}{Numerical Analyst:} For each room, Bob needs to wash 2 sheets, 1 comforter, 4 pillow cases (since $2 \times 2 = 4$), and 8 towels (since $4 \times 2 = 8$). Therefore, for each room, the total amount of laundry Bob needs to wash is:\\
\(2 + 1 + 4 + 8 = 15\) pieces of laundry.\\
For 80 rooms, the total number of pieces of laundry Bob must wash is:\\
\(80 \times 15 = 1200\) pieces of laundry.\\
Thus, the final answer is: \boxed{1200}.\\

\colorbox{Yellow}{Judge:} The candidate correctly interprets the problem's relationships: sheets: 2 (given), comforter: 1 (given), pillow cases: twice the sheets, \(2 \times 2 = 4\) (correct), towels: twice the pillow cases, \(2 \times 4 = 8\) (correct). Total per room: \(2 + 1 + 4 + 8 = 15\) (correct). Total for 80 rooms: \(80 \times 15 = 1200\) (correct). All steps align with the problem's specifications, and the final answer is accurate. The candidate’s solution is error-free.\\

Score: 1.\\

\colorbox{Red}{Mathematician:} Every room is equipped with 2 sheets, 1 comforter, 4 pillow cases (calculated as \(2 \times 2\)), and 8 towels (since \(4 \times 2 = 8\)). Summing these, the total number of laundry items for one room is:\\
\(2 + 4 + 8 = 14\) pieces of laundry.\\
Multiplying by the number of rooms, the total quantity for 80 rooms becomes:\\
\(80 \times 14 = \boxed{1120}\).\\

\colorbox{Yellow}{Judge:} The candidate incorrectly omitted the comforter in their final sum. Here’s the breakdown: sheets: 2 (correct), comforter: 1 (given but not included in the candidate’s calculation), pillow cases: \(2 \times 2 = 4\) (correct), towels: \(2 \times 4 = 8\) (correct). The candidate’s total per room was \(2 + 4 + 8 = 14\), but the correct total should include the comforter: \(2 + 1 + 4 + 8 = 15\). Thus, the total for 80 rooms should be \(80 \times 15 = 1200\), not 1120. The error stems from excluding the comforter in the summation.\\

Score: 0.\\

\colorbox{lightgray}{$\mathcal{D}_{S_1}$ rank:} [\textcolor{ForestGreen}{Numerical Analyst}, \textcolor{red}{Mathematical Modeler}, \textcolor{red}{Mathematician}]\\

\noindent\hfill\makebox[0.6\linewidth]{\textbf{Iteration 2}}\hfill\null
\medskip 

\colorbox{OliveGreen}{Mathematical Modeler:}
Each room contains 2 sheets, 1 comforter, 2 pillow cases (which is twice the number of sheets), and 4 towels (which is twice the number of pillow cases). Hence, the total number of laundry items in each room is:\\
\(2 + 1 + 4 + 8 = 15\) pieces of laundry.\\
When considering 80 rooms, the total number of pieces of laundry is:\\
\(80 \times 15 = 1200\) pieces of laundry.\\
Thus, the final answer is: \boxed{1200}.

This confirms the accuracy of the revised calculation.\\

\colorbox{Yellow}{Judge:} The candidate’s final answer is correct, and the mathematical calculations align with the problem’s requirements: 1. Sheets: 2 (correct). 2. Comforter: 1 (included in the total). 3. Pillow Cases: The problem states ``twice as many pillow cases as sheets.'' Since there are 2 sheets, this should be \(2 \times 2 = 4\). The candidate correctly uses 4 pillow cases in their total calculation (despite a typographical inconsistency in the explanation, where they initially wrote ``2 pillow cases''). 4. Towels: ``Twice as many towels as pillow cases'' gives \(2 \times 4 = 8\) (correct).\\

The candidate’s total per room is \(2 + 1 + 4 + 8 = 15\) (correct), leading to \(80 \times 15 = 1200\) (correct).\\

While the explanation contains a minor inconsistency in wording (``2 pillow cases'' instead of 4), the numerical calculations are accurate. Since the final answer is correct and the mathematical steps resolve the inconsistency, the solution is valid.\\

Score: 1.\\

\colorbox{OliveGreen}{Numerical Analyst:} 
Based on the information provided, we know that each room includes 2 sheets, 1 comforter, 4 pillow cases (computed as \(2 \times 2\)), and 8 towels (since \(4 \times 2 = 8\)). Adding these together, the total number of laundry items required for one room is:\\
\(2 + 1 + 4 + 8 = 15\) pieces.\\
Multiplying by the total number of rooms, the overall amount of laundry for 80 rooms is:\\
\(80 \times 15 = \boxed{1200}\).\\

Although the other agents correctly determined the number of laundry items required for each room, there was an error in the concluding step--calculate the total number of laundry pieces.\\

\colorbox{Yellow}{Judge:} The candidate’s calculations and final answer are correct: sheets: 2 (given), comforter: 1 (given), pillow cases: \(2 \times 2 = 4\) (twice the sheets), towels: \(2 \times 4 = 8\) (twice the pillow cases), total per room: \(2 + 1 + 4 + 8 = 15\) (correct), total for 80 rooms: \(80 \times 15 = 1200\) (correct).\\

Score: 1.\\

\colorbox{Red}{Mathematician:}  According to the provided details, every room is equipped with 2 sheets, 1 comforter, 4 pillow cases (calculated as \(2 \times 2\)), and 8 towels (since \(4 \times 2 = 8\)). Summing these, the total number of laundry items for one room is:\\
\(1 + 4 + 8 = 13\) pieces of laundry.\\
Multiplying by the number of rooms, the total quantity for 80 rooms becomes:\\
\(80 \times 13 = \boxed{1040}\).\\

\colorbox{Yellow}{Judge:} The candidate incorrectly omitted the sheets in their final sum. Here’s the breakdown: 1. Sheets: 2 (given but not included in the candidate’s calculation). 2. Comforter: 1 (included). 3. Pillow Cases: \(2 \times 2 = 4\) (correct). 4. Towels: \(2 \times 4 = 8\) (correct). The candidate’s total per room was \(1 + 4 + 8 = 13\), but the correct total should include the sheets: \(2 + 1 + 4 + 8 = 15\). Thus, the total for 80 rooms should be \(80 \times 15 = 1200\), not \(1040\). The error stems from excluding the sheets in the summation.\\

Score: 0.\\

\colorbox{lightgray}{$\mathcal{D}_{S_2}$ rank:} [\textcolor{ForestGreen}{Mathematical Modeler}, \textcolor{ForestGreen}{Numerical Analyst}, \textcolor{red}{Mathematician}]\\

\noindent\hfill\makebox[0.6\linewidth]{\textbf{Iteration 3}}\hfill\null
\medskip 

\colorbox{OliveGreen}{Mathematical Modeler:} From the details provided, it is clear that each room contains 2 sheets, 1 comforter, 4 pillow cases (as \(2 \times 2 = 4\)), and 8 towels (calculated as \(4 \times 2 = 8\)). Adding these together, the total number of laundry items per room amounts to:\\
\(2 + 1 + 4 + 8 = 15\) items.\\
Consequently, for 80 rooms, the total number of laundry pieces becomes:\\
\(80 \times 15 = \boxed{1200}\).\\

I agree with the other agent's corrected 
calculation.\\

\colorbox{Yellow}{Judge:} The candidate’s solution is mathematically accurate and complete: sheets: 2 (correct), comforter: 1 (correct), pillow cases: \(2 \times 2 = 4\) (twice the sheets), towels: \(2 \times 4 = 8\) (twice the pillow cases), total per room: \(2 + 1 + 4 + 8 = 15\) (correct), total for 80 rooms: \(80 \times 15 = 1200\) (correct). \\

Score: 1.\\

\colorbox{OliveGreen}{Numerical Analyst:} 
Given the information, we know that each room includes 2 sheets, 1 comforter, 4 pillow cases (computed as \(2 \times 2\)), and 8 towels (since \(4 \times 2 = 8\)). Adding these together, the total number of laundry items required for one room is:\\
\(2 + 1 + 4 + 8 = 15\) pieces.\\
Multiplying by the total number of rooms, the overall amount of laundry for 80 rooms is:\\
\(80 \times 15 = \boxed{1200}\).\\

Although the other agent correctly determined the number of laundry items required for each room, there was an error in the concluding step. Specifically, they incorrectly used \(80 \times 13\) instead of the appropriate multiplication of \(80 \times 15\) to calculate the total number of laundry pieces.\\

\colorbox{Yellow}{Judge:} The candidate’s solution is fully correct: sheets: 2 (given), comforter: 1 (given), pillow cases: \(2 \times 2 = 4\) (twice the sheets), towels: \(2 \times 4 = 8\) (twice the pillow cases), total per room: \(2 + 1 + 4 + 8 = 15\) (correct), total for 80 rooms: \(80 \times 15 = 1200\) (correct). All steps align with the problem’s specifications, and the final answer is accurate. No errors were made.\\

Score: 1.\\

\colorbox{OliveGreen}{Mathematician:}
According to the given details, each room contains 2 sheets, 1 comforter, 4 pillow cases (calculated as \(2 \times 2\)), and 8 towels (determined by \(4 \times 2\)). Summing these amounts, the total number of laundry items per room is:\\
\(2 + 1 + 4 + 8 = 15\) pieces of laundry.\\
To find the total for 80 rooms, we multiply:\\
\(80 \times 15 = \boxed{1200}\).\\

I concur with the other agents’ results and the computations they presented.\\

\colorbox{Yellow}{Judge:} The candidate’s solution is fully correct and aligns with the problem’s requirements: sheets: 2 (given), comforter: 1 (given), pillow cases: \(2 \times 2 = 4\) (twice the sheets), towels: \(2 \times 4 = 8\) (twice the pillow cases), total per room: \(2 + 1 + 4 + 8 = 15\) (correct), total for 80 rooms: \(80 \times 15 = 1200\) (correct). All calculations are accurate, and the final answer matches the problem’s constraints.\\

Score: 1.\\

\colorbox{lightgray}{$\mathcal{D}_{S_3}$ rank:} [\textcolor{ForestGreen}{Mathematician}, \textcolor{ForestGreen}{Mathematical Modeler}, \textcolor{ForestGreen}{Numerical Analyst}]\\
\end{tcolorbox}

\begin{tcolorbox}[breakable, title=Example for \textsc{Correct Coding Task \colorbox{OliveGreen}{Green}: Correct Solution. \colorbox{Red}{Red}: Wrong Solution.}, colback=blue!5!white, colframe=blue!25!black, fonttitle=\footnotesize, boxrule=0.2mm, sharp corners]
\label{}
\small

\colorbox{pink}{Query:} Yarik is a big fan of many kinds of music. But Yarik loves not only listening to music but also writing it. He likes electronic music most of all, so he has created his own system of music notes, which, in his opinion, is best for it.\\

Since Yarik also likes informatics, in his system notes are denoted by integers of $2^k$, where $k \ge 1$ — a positive integer. But, as you know, you can't use just notes to write music, so Yarik uses combinations of two notes. The combination of two notes $(a, b)$, where $a = 2^k$ and $b = 2^l$, he denotes by the integer $a^b$.\\

For example, if $a = 8 = 2^3$, $b = 4 = 2^2$, then the combination $(a, b)$ is denoted by the integer $a^b = 8^4 = 4096$. Note that different combinations can have the same notation, e.g., the combination $(64, 2)$ is also denoted by the integer $4096 = 64^2$.\\

Yarik has already chosen $n$ notes that he wants to use in his new melody. However, since their integers can be very large, he has written them down as an array $a$ of length $n$, then the note $i$ is $b_i = 2^{a_i}$. The integers in array $a$ can be repeated.\\

The melody will consist of several combinations of two notes. Yarik was wondering how many pairs of notes $b_i, b_j$ $(i < j)$ exist such that the combination $(b_i, b_j)$ is equal to the combination $(b_j, b_i)$. In other words, he wants to count the number of pairs $(i, j)$ $(i < j)$ such that $b_i^{b_j} = b_j^{b_i}$. Help him find the number of such pairs.\\

Input\\

The first line of the input contains one integer $t$ ($1 \le t \le 10^4$) — the number of test cases.\\

The first line of each test case contains one integer $n$ ($1 \leq n \leq 2 \cdot 10^5$) — the length of the arrays.\\

The next line contains $n$ integers $a_1, a_2, \dots, a_n$ ($1 \leq a_i \leq 10^9$) — array $a$.\\

It is guaranteed that the sum of $n$ over all test cases does not exceed $2 \cdot 10^5$.\\

Output\\

For each test case, output the number of pairs that satisfy the given condition.\\

Sample Input 1:\\

5\\

1\\

2\\

4\\

3 1 3 2\\

2\\

1000 1000\\

3\\

1 1 1\\

19\\

2 4 1 6 2 8 5 4 2 10 5 10 8 7 4 3 2 6 10\\

Sample Output 1:\\

0\\
2\\
1\\
3\\
19\\

Make sure to wrap your code in \texttt{\textasciigrave\textasciigrave\textasciigrave python \textasciigrave\textasciigrave\textasciigrave} block and Markdown delimiters, and include exactly one block of code with the entire solution (in the final code step).\\

\noindent\hfill\makebox[0.6\linewidth]{\textbf{Iteration 1}}\hfill\null
\medskip 

\colorbox{Red}{Implementation Specialist:} To solve this problem, we need to determine the number of pairs of notes $(b_i, b_j)$ such that $b_i^{b_j} = b_j^{b_i}$. The notes are given as $b_i = 2^{a_i}$, where $a_i$ is an integer. The key insight is to recognize the conditions under which the equation $b_i^{b_j} = b_j^{b_i}$ holds. This equation can be broken down into two main cases. The first case is when the values are the same, meaning $a_i = a_j$. In this situation, the equation trivially holds because both sides will be identical. The number of such pairs is given by the combination formula $C(n, 2) = \frac{n(n-1)}{2}$ for each group of identical values. The second case occurs when the values are different but satisfy a special condition: the equation holds when one value is 1 and the other is 2. This is derived from the mathematical simplification of the equation $a_i \cdot 2^{a_j} = a_j \cdot 2^{a_i}$, which only holds true for $a_i = 1$ and $a_j = 2$ (or vice versa) when $a_i \neq a_j$.\\

\texttt{\textasciigrave\textasciigrave\textasciigrave python}
\begin{verbatim}
import sys
from collections import Counter

def main():
    # Read all input tokens at once
    x = sys.stdin.read().split()
    i = 0  # Index to track position in input list
    
    t = int(x[i])  # Number of test cases
    i += 1
    
    for _ in range(t):
        n = int(x[i])  # Number of elements in this test case
        i += 1
        
        # Extract the list of integers for current test case
        l = list(map(int, x[i:i+n]))
        i += n
        
        # Count frequencies of each element
        c = Counter(l)
        
        s = 0  # Sum of pairs
        
        # Sum pairs of identical elements: count choose 2
        for v in c.values():
            s += v * (v - 1) // 2
        
        # Add pairs formed by one element '1' and one element '2'
        s += c.get(1, 0) * c.get(2, 0)
        
        print(s)

if __name__ == "__main__":
    main()
\end{verbatim}
\texttt{\textasciigrave\textasciigrave\textasciigrave}\\

\colorbox{Yellow}{Judge:} There are 10 test cases. The candidate passes 5 test cases.\\

Score: 5.\\

\colorbox{Red}{Programmer:} To solve this problem, we need to determine the number of pairs of notes such that the combination of two notes $(b_i, b_j)$ results in the same integer as the combination $(b_j, b_i)$. The notes are given as $b_i = 2^{a_i}$, and we are looking for pairs $(i, j)$ where $b_i^{b_j} = b_j^{b_i}$. The condition $b_i^{b_j} = b_j^{b_i}$ can be simplified to checking if $a_i \cdot 2^{a_j} = a_j \cdot 2^{a_i}$, which holds true in two scenarios: first, when $a_i = a_j$, producing identical pairs; and second, when one value is 1 and the other is 2, creating valid cross pairs. To count the identical pairs, for each unique value $x$ in the array, we use the combination formula $\binom{count}{2}$, where $count$ is the frequency of $x$. For cross pairs, which occur specifically between values 1 and 2, the number of such pairs is calculated as the product of the counts of 1 and 2.\\

\texttt{\textasciigrave\textasciigrave\textasciigrave python}
\begin{verbatim}
import sys
from collections import defaultdict

def main():
    x = sys.stdin.read().split()  # Read all input tokens
    i = 0  # Input index pointer
    
    t = int(x[i])  # Number of test cases
    i += 1
    
    for _ in range(t):
        n = int(x[i])  # Number of elements
        i += 1
        
        l = list(map(int, x[i:i+n]))  # List of integers for this test case
        i += n
        
        f = defaultdict(int)  # Frequency dictionary
        
        # Count occurrences of each number
        for v in l:
            f[v] += 1
        
        s = 0  # Count pairs of identical elements
        
        # Sum pairs from frequencies (nC2)
        for v in f.values():
            s += v * (v - 1) // 2
        
        # Count pairs formed by one '1' and one '2'
        c = f.get(1, 0) * f.get(2, 0)
        
        print(s + c)

if __name__ == "__main__":
    main()
\end{verbatim}
\texttt{\textasciigrave\textasciigrave\textasciigrave}\\

\colorbox{Yellow}{Judge:} There are 10 test cases. The candidate passes 5 test cases.\\

Score: 5.\\

\colorbox{Red}{Data Structure Specialist:} To solve this problem, we need to determine the number of pairs of music notes, represented by integers of the form $2^k$, such that the combination of two notes $(b_i, b_j)$ results in the same value as $(b_j, b_i)$. In other words, we seek pairs $(i, j)$ where $b_i^{b_j} = b_j^{b_i}$. By applying logarithms and using the properties of exponents, for notes defined as $b_i = 2^{a_i}$ and $b_j = 2^{a_j}$, the condition simplifies to $a_i \cdot 2^{a_j} = a_j \cdot 2^{a_i}$. Two key observations arise from this: first, if $a_i = a_j$, the equation holds true, meaning all pairs of identical values satisfy the condition; second, the equation also holds when one value is 1 and the other is 2, since $1 \cdot 2^2 = 2 \cdot 2^1$ simplifies to $4 = 4$. The algorithm to find the number of such pairs involves two parts. For pairs with the same value $v$, we count the number of pairs using the combination formula $C(n, 2) = \frac{n(n-1)}{2}$, where $n$ is the frequency of $v$. For cross-value pairs where one note corresponds to 1 and the other to 2, the number of valid pairs is simply the product of their counts.\\

\texttt{\textasciigrave\textasciigrave\textasciigrave python}
\begin{verbatim}
import sys
from collections import Counter

def main():
    x = sys.stdin.read().split()  # Read all input tokens at once
    i = 0  # Input index pointer
    
    t = int(x[i])  # Number of test cases
    i += 1
    
    for _ in range(t):
        n = int(x[i])  # Number of elements
        i += 1
        
        l = list(map(int, x[i:i+n]))  # List of integers for this test case
        i += n
        
        c = Counter(l)  # Frequency counter
        
        c1 = c.get(1, 0)  # Count of 1's
        c2 = c.get(2, 0)  # Count of 2's
        
        x = 0
        
        # Combine counts of 1 and 2, calculate pairs among them
        m = c1 + c2
        x += m * (m - 1) // 2
        
        # Calculate pairs for other numbers except 1 and 2
        for k in c:
            if k not in (1, 2):
                v = c[k]
                x += v * (v - 1) // 2
        
        print(x)

if __name__ == "__main__":
    main()
\end{verbatim}
\texttt{\textasciigrave\textasciigrave\textasciigrave}\\

\colorbox{Yellow}{Judge:} There are 10 test cases. The candidate passes 5 test cases.\\

Score: 5.\\

\colorbox{lightgray}{$\mathcal{D}_{S_1}$ rank:} [\textcolor{red}{Implementation Specialist}, \textcolor{red}{Programmer}, \textcolor{red}{Data Structure Specialist}]\\

\noindent\hfill\makebox[0.6\linewidth]{\textbf{Iteration 2}}\hfill\null
\medskip 

\colorbox{Red}{Implementation Specialist:} : After reviewing the solutions 
provided by other agents, I agree that to solve this problem, we need to determine the number of pairs of notes $(b_i, b_j)$ such that $b_i^{b_j} = b_j^{b_i}$, where each note is given by $b_i = 2^{a_i}$. We are interested in counting pairs $(i, j)$ with $i < j$ that satisfy this condition. By applying properties of exponents and substituting $b_i = 2^{a_i}$ and $b_j = 2^{a_j}$, the condition simplifies to checking when $\frac{2^{a_i}}{a_i} = \frac{2^{a_j}}{a_j}$. There are two key observations: first, if $a_i = a_j$, the condition is automatically met; second, pairs where one value is 1 and the other is 2 also satisfy the condition, since $\frac{2^1}{1} = 2$ and $\frac{2^2}{2} = 2$. The algorithm involves counting the frequency of each $a_i$, then calculating the number of valid pairs for identical values using the combination formula $\binom{f}{2}$, where $f$ is the frequency of the value. Finally, to account for the special case of pairs between 1 and 2, we add the product of their counts to the total.\\

\texttt{\textasciigrave\textasciigrave\textasciigrave python}
\begin{verbatim}
import sys
from collections import Counter

def main():
    x = sys.stdin.read().split()  # Read all input tokens at once
    i = 0  # Index pointer
    
    t = int(x[i])  # Number of test cases
    i += 1
    
    for _ in range(t):
        n = int(x[i])  # Number of elements in this test case
        i += 1
        
        l = list(map(int, x[i:i+n]))  # List of elements
        i += n
        
        c = Counter(l)  # Frequency count
        
        s = 0  # Sum of pairs
        
        # Calculate pairs from identical elements: nC2 for each count
        for v in c.values():
            s += v * (v - 1) // 2
        
        # Add pairs formed by one '1' and one '2'
        s += c.get(1, 0) * c.get(2, 0)
        
        print(s)

if __name__ == "__main__":
    main()
\end{verbatim}
\texttt{\textasciigrave\textasciigrave\textasciigrave}\\

\colorbox{Yellow}{Judge:} There are 10 test cases. The candidate passes 6 test cases.\\

Score: 6.\\

\colorbox{Red}{Programmer:} I agree with the other agents that, to solve this problem, we need to determine the number of valid pairs of notes such that the combination of two notes $(b_i, b_j)$ results in the same integer as the combination $(b_j, b_i)$. This is equivalent to finding pairs $(i, j)$ where $b_i^{b_j} = b_j^{b_i}$. The key insight is that this equation holds under two conditions: first, when the exponents $a_i$ and $a_j$ are equal, the pairs $(b_i, b_j)$ will always satisfy the equation; second, when one exponent is 1 and the other is 2, the equation also holds. To solve the problem, the approach involves three steps. First, use a frequency counter to count the occurrences of each value in the array. Second, for each value $x$, compute the number of valid pairs among identical values using the combination formula $\binom{f}{2}$, where $f$ is the frequency of $x$. Third, calculate the number of specific pairs where one value is 1 and the other is 2 by multiplying their counts.\\

\texttt{\textasciigrave\textasciigrave\textasciigrave python}
\begin{verbatim}
import sys
from collections import Counter

def main():
    t = int(sys.stdin.readline())  # Number of test cases
    
    for _ in range(t):
        n = int(sys.stdin.readline().strip())  # Number of elements
        l = list(map(int, sys.stdin.readline().split()))  # List of integers
        
        c = Counter(l)  # Frequency counter
        
        s = 0  # Sum of pairs
        
        # Calculate pairs among identical elements: nC2 for each frequency
        for v in c.values():
            s += v * (v - 1) // 2
        
        # Add pairs formed by one '1' and one '2'
        s += c.get(1, 0) * c.get(2, 0)
        
        print(s)

if __name__ == "__main__":
    main()
\end{verbatim}
\texttt{\textasciigrave\textasciigrave\textasciigrave}\\

\colorbox{Yellow}{Judge:} There are 10 test cases. The candidate passes 5 test cases.\\

Score: 5.\\

\colorbox{OliveGreen}{Data Structure Specialist:} To solve this problem, we need to determine the number of valid pairs of notes such that the combination of two notes $(b_i, b_j)$ results in the same integer as the combination $(b_j, b_i)$. This is equivalent to finding pairs $(i, j)$ where $b_i^{b_j} = b_j^{b_i}$. The key insight comes from the equation $b_i^{b_j} = b_j^{b_i}$, where $b_i = 2^{a_i}$ and $b_j = 2^{a_j}$. This equation simplifies to $a_i \cdot 2^{a_j} = a_j \cdot 2^{a_i}$. To solve this efficiently, we follow several steps. First, for each $a_i$, we perform prime factorization, expressing it as $2^y \times x$, where $x$ is the odd part of $a_i$ and $y$ is the exponent of 2 in its factorization. Next, we generate a key for each $a_i$ in the form $(x, y - a_i)$. This key is designed such that if two numbers $a_i$ and $a_j$ share the same key, they satisfy the equation $a_i \cdot 2^{a_j} = a_j \cdot 2^{a_i}$. We then use a frequency dictionary to count the occurrences of each key. Finally, for each key, we compute the number of valid pairs using the combination formula $\binom{f}{2}$, where $f$ is the frequency of that key.\\

\texttt{\textasciigrave\textasciigrave\textasciigrave python}
\begin{verbatim}
import sys
from collections import defaultdict

def main():
    x = sys.stdin.read().split()  # Read all input tokens at once
    i = 0  # Input index pointer
    
    t = int(x[i])  # Number of test cases
    i += 1
    
    for _ in range(t):
        n = int(x[i])  # Number of elements in this test case
        i += 1
        
        l = list(map(int, x[i:i+n]))  # List of integers
        i += n
        
        f = defaultdict(int)  # Frequency dictionary
        
        for v in l:
            a = v
            b = 0
            # Count how many times 2 divides the number
            while a % 2 == 0:
                a //= 2
                b += 1
            # Key is tuple of (odd part, difference of exponents)
            k = (a, b - v)
            f[k] += 1
        
        s = 0  # Result accumulator
        
        # Count pairs for each unique key using nC2 formula
        for c in f.values():
            s += c * (c - 1) // 2
        
        print(s)

if __name__ == "__main__":
    main()
\end{verbatim}
\texttt{\textasciigrave\textasciigrave\textasciigrave}\\

\colorbox{Yellow}{Judge:} There are 10 test cases. The candidate passes all 10 test cases with correct result.\\

Score: 10.\\

\colorbox{lightgray}{$\mathcal{D}_{S_2}$ rank:} [\textcolor{ForestGreen}{Data Structure Specialist}, \textcolor{red}{Implementation Specialist}, \textcolor{red}{Programmer}]\\

\noindent\hfill\makebox[0.6\linewidth]{\textbf{Iteration 3}}\hfill\null
\medskip 

\colorbox{OliveGreen}{Implementation Specialist:} I concur with the other agent that to tackle this problem, our goal is to find the number of valid pairs of notes $(b_i, b_j)$ such that raising one note to the power of the other yields the same result when the order is reversed; that is, pairs $(i, j)$ satisfying $b_i^{b_j} = b_j^{b_i}$. Given that each note can be expressed as $b_i = 2^{a_i}$, the problem reduces to analyzing the equation $a_i \cdot 2^{a_j} = a_j \cdot 2^{a_i}$. To approach this effectively, we begin by decomposing each exponent $a_i$ into its prime factors, specifically writing it as $2^y \times x$, where $x$ is the odd component and $y$ indicates the power of two within the factorization. Using this factorization, we create a unique key for each $a_i$ defined by the pair $(x, y - a_i)$. This construction ensures that any two exponents sharing the same key satisfy the original equation. After generating these keys for all elements, we count their frequencies using a dictionary. Finally, to determine the number of valid pairs, we apply the combination formula $\binom{f}{2}$ for each key, where $f$ is the frequency of that key, summing these results to obtain the total count of valid pairs.\\

\texttt{\textasciigrave\textasciigrave\textasciigrave python}
\begin{verbatim}
import sys
from collections import defaultdict

def main():
    d = sys.stdin.read().strip().split()  # Read and split all input
    t = int(d[0])  # Number of test cases
    p = 1  # Current position in input list
    
    for _ in range(t):
        n = int(d[p])  # Number of elements in this test case
        p += 1
        
        l = list(map(int, d[p:p+n]))  # Current list of elements
        p += n
        
        f = defaultdict(int)  # Frequency dictionary
        
        for v in l:
            o = v  # Original value
            c = 0  # Count of factors 2
            
            # Count how many times 2 divides the value
            while v % 2 == 0:
                v //= 2
                c += 1
            
            # Key: (odd part, count_twos - original_value)
            k = (v, c - o)
            f[k] += 1
        
        s = 0  # Result accumulator
        
        # Calculate pairs for each unique key using nC2
        for x in f.values():
            s += x * (x - 1) // 2
        
        print(s)

if __name__ == "__main__":
    main()
\end{verbatim}
\texttt{\textasciigrave\textasciigrave\textasciigrave}\\

\colorbox{Yellow}{Judge:} There are 10 test cases. The candidate passes all 10 test cases with correct result.\\

Score: 10.\\

\colorbox{OliveGreen}{Programmer:} I agree with the other agent that to address this problem, we aim to find the number of valid pairs of notes $(b_i, b_j)$ such that the expression $b_i^{b_j}$ equals $b_j^{b_i}$. In other words, we seek pairs $(i, j)$ where the exponentiation relation $b_i^{b_j} = b_j^{b_i}$ holds true. Since each note is defined as $b_i = 2^{a_i}$, the equation simplifies to $a_i \cdot 2^{a_j} = a_j \cdot 2^{a_i}$. To efficiently solve this, we first factorize each exponent $a_i$ into the form $2^y \times x$, where $x$ represents the odd component and $y$ is the power of two in its prime factorization. Using this decomposition, we construct a key for each $a_i$ as $(x, y - a_i)$. This key is crafted so that if two exponents $a_i$ and $a_j$ share the same key, they satisfy the equality condition. Next, we tally the frequency of each key using a dictionary or hash map. Finally, the number of valid pairs corresponding to each key is computed using the combination formula $\binom{f}{2}$, where $f$ denotes the frequency of that key. Summing these values yields the total count of valid pairs.\\

\texttt{\textasciigrave\textasciigrave\textasciigrave python}
\begin{verbatim}
import sys
from collections import defaultdict

def main():
    d = sys.stdin.read().strip().split()  # Read all input tokens
    t = int(d[0])  # Number of test cases
    p = 1  # Position pointer in input list
    
    for _ in range(t):
        n = int(d[p])  # Number of elements
        p += 1
        
        l = list(map(int, d[p:p+n]))  # List of elements
        p += n
        
        f = defaultdict(int)  # Frequency dictionary
        
        for v in l:
            b = v  # Base value
            c = 0  # Count of factors 2
            
            # Count how many times 2 divides b
            while b % 2 == 0:
                b //= 2
                c += 1
            
            # Key is tuple: (odd part, power difference)
            k = (b, c - v)
            f[k] += 1
        
        s = 0  # Result accumulator
        
        # Count pairs for each frequency using nC2 formula
        for x in f.values():
            s += x * (x - 1) // 2
        
        print(s)

if __name__ == "__main__":
    main()
\end{verbatim}
\texttt{\textasciigrave\textasciigrave\textasciigrave}\\

\colorbox{Yellow}{Judge:} There are 10 test cases. The candidate passes all 10 test cases with correct result.\\

Score: 10.\\

\colorbox{OliveGreen}{Data Structure Specialist:}
To solve this problem, the goal is to identify the number of valid pairs of notes $(b_i, b_j)$ for which the value of $b_i^{b_j}$ equals that of $b_j^{b_i}$. This translates to finding pairs $(i, j)$ satisfying the equation $b_i^{b_j} = b_j^{b_i}$. Given that each note $b_i$ can be expressed as $2^{a_i}$, the equation simplifies to $a_i \cdot 2^{a_j} = a_j \cdot 2^{a_i}$. To approach this efficiently, each exponent $a_i$ is factorized into the form $2^y \times x$, where $x$ denotes the odd part of $a_i$ and $y$ is the exponent of 2 in its prime factorization. Based on this factorization, we construct a unique key $(x, y - a_i)$ for each $a_i$. This key is carefully designed so that two numbers sharing the same key satisfy the given equality condition. Once the keys are generated, a frequency map is used to count how many times each key appears. Finally, for each key, the number of valid pairs is calculated using the combination formula $\binom{f}{2}$, where $f$ represents the frequency of that key. Summing these counts across all keys gives the total number of valid pairs.\\

\texttt{\textasciigrave\textasciigrave\textasciigrave python}
\begin{verbatim}
import sys
from collections import defaultdict

def main():
    d = sys.stdin.read().split()  # Read all input tokens
    i = 1  # Index pointer after reading t
    
    t = int(d[0])  # Number of test cases
    
    for _ in range(t):
        n = int(d[i])  # Number of elements
        i += 1
        
        l = list(map(int, d[i:i+n]))  # Current list of elements
        i += n
        
        f = defaultdict(int)  # Frequency dictionary
        
        for v in l:
            x = v  # Current value
            p = 0  # Count of factor 2
            
            # Count how many times 2 divides x
            while x % 2 == 0:
                x //= 2
                p += 1
            
            k = (x, p)  # Key as (odd part, power of two)
            f[k] += 1
        
        s = 0  # Result accumulator
        
        # Count pairs for each unique key using nC2 formula
        for c in f.values():
            s += c * (c - 1) // 2
        
        print(s)

if __name__ == "__main__":
    main()
\end{verbatim}
\texttt{\textasciigrave\textasciigrave\textasciigrave}\\

While the previous codes are correct for the given problem constraints, it implicitly assumes that only pairs of equal values or (1, 2) satisfy the equation later searching iteratively. However, explicitly grouping all valid pairs using a mathematical key, making it more robust and efficient for all scenarios, including possible edge cases not covered by the first approach.\\

\colorbox{Yellow}{Judge:} There are 10 test cases. The candidate passes all 10 test cases with correct result.\\

Score: 10.\\

\colorbox{lightgray}{$\mathcal{D}_{S_3}$ rank:} [\textcolor{ForestGreen}{Data Structure Specialist}, \textcolor{ForestGreen}{Programmer}, \textcolor{ForestGreen}{Implementation Specialist}]\\

\end{tcolorbox}

\section{External Retrieval Corpus for Code}
\label{sec:appendix:ext_retrieve}
\begin{figure}[h]
    \centering
    \includegraphics[width=\linewidth]{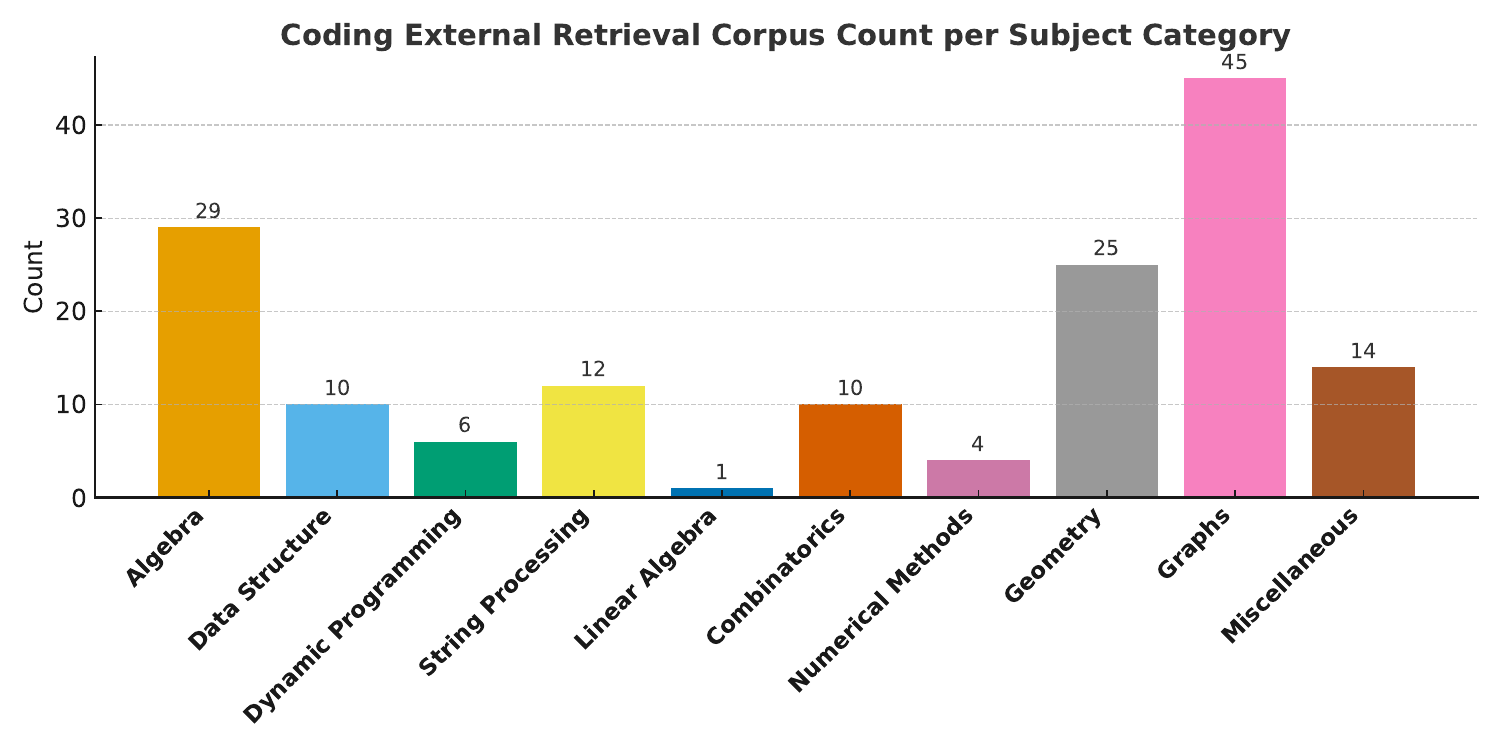}
    \caption{Coding External Retrieval Corpus Count per Subject Category.}
    \label{fig:coding_external_retrieval_corpus_count}
\end{figure}

As shown in Figure \ref{fig:coding_external_retrieval_corpus_count}, our coding external retrieval corpus has 156 fine grained algorithmic tutorials with relevant problems along with their solutions collections from 10 different categories (algebrea, data structure, dynamic programming, string processing, linear algebrea, combinatorics, numerical methods, geometry, graphs and miscellaneous).

\section{Additional Analysis}
This section presents our multiple runs' results which we tested and list of roles which LLMs selected during the experiments.

\subsection{Patterns in \xolver Reasoning Traces}
\label{sec:appendix:reasoning}
We employ the QWQ-32B model for the initial segmentation of the chain of thought, guided by the first prompt illustrated in Appendix \ref{sec:appendix:reaso_seg}. Subsequently, for segments that remain unannotated, a second segmentation pass is performed using the second prompt depicted in Appendix \ref{sec:appendix:reaso_seg}. Reasoning patterns are identified by extracting content enclosed within the <pattern> tags. Verification and reasoning patterns are combined into a unified self-evaluation category. If the model assigns multiple patterns to a single segment, that segment is excluded due to ambiguity in pattern classification. For each generated output, we calculate the proportion of occurrences of each pattern relative to the total patterns present, resulting in a frequency vector representing pattern distribution per generation. In examining the relationship between pattern usage and problem difficulty, we compute the mean frequencies separately for correct and incorrect generations and assess significance through a t-test. To evaluate pattern prevalence on a per-problem basis, a binary matrix is constructed where rows correspond to problems and entries indicate whether a pattern is more common in correct (1) or incorrect (0) solutions. The statistical significance of these findings is evaluated using a binomial test.

\begin{table}[h]
\centering
\adjustbox{max width=\textwidth}{
\begin{tabular}{lcccc}
\hline
\textbf{Reasoning Pattern} 
& \multicolumn{2}{c}{\textbf{Correct Solutions}} 
& \multicolumn{2}{c}{\textbf{Incorrect Solutions}} \\
& Easy $\rightarrow$ Medium & Medium $\rightarrow$ High 
& Easy $\rightarrow$ Medium & Medium $\rightarrow$ High \\
\hline
\multicolumn{5}{l}{\textbf{(a) OpenCodeReasoning}} \\
\hline
Self-Evaluation ($\downarrow$)  
& \highlightcell{\textcolor{red}{0.39 $\rightarrow$ 0.37}} & \highlightcell{\textcolor{red}{0.37 $\rightarrow$ 0.34}} 
& 0.36 $\rightarrow$ 0.37 & \highlightcell{\textcolor{red}{0.34 $\rightarrow$ 0.31}} \\
New Approach ($\uparrow$)
& {\textcolor{OliveGreen}{0.16 $\rightarrow$ 0.20}} & {\textcolor{OliveGreen}{0.20 $\rightarrow$ 0.23}} 
& {\textcolor{OliveGreen}{0.16 $\rightarrow$ 0.22}} & {\textcolor{OliveGreen}{0.22 $\rightarrow$ 0.25}} \\
Problem Rephrasing ($\downarrow\uparrow$)
& \textcolor{red}{0.21 $\rightarrow$ 0.20} & 0.20 $\rightarrow$ 0.20 
& 0.21 $\rightarrow$ 0.22 & \textcolor{OliveGreen}{0.22 $\rightarrow$ 0.23} \\
Subgoal Setup ($\downarrow$)
& \textcolor{red}{0.13 $\rightarrow$ 0.12} & \textcolor{red}{0.12 $\rightarrow$ 0.10} 
& \highlightcell{\textcolor{red}{0.13 $\rightarrow$ 0.10}} & 0.10 $\rightarrow$ 0.10 \\
\hline
\multicolumn{5}{l}{\textbf{(b) \method{\xolver}}} \\
\hline
Self-Evaluation ($\uparrow$)           
& \highlightcell{\textcolor{OliveGreen}{0.35 $\rightarrow$ 0.38}} & \highlightcell{\textcolor{OliveGreen}{0.38 $\rightarrow$ 0.40}} 
& 0.35 $\rightarrow$ 0.37 & \highlightcell{\textcolor{OliveGreen}{0.32 $\rightarrow$ 0.35}} \\
New Approach ($\uparrow$)
& {\textcolor{OliveGreen}{0.18 $\rightarrow$ 0.21}} & {\textcolor{OliveGreen}{0.21 $\rightarrow$ 0.24}} 
& {\textcolor{OliveGreen}{0.17 $\rightarrow$ 0.24}} & {\textcolor{OliveGreen}{0.24 $\rightarrow$ 0.26}} \\
Problem Rephrasing ($\downarrow\uparrow$)         
& \textcolor{red}{0.20 $\rightarrow$ 0.17} & 0.18 $\rightarrow$ 0.18
& 0.23 $\rightarrow$ 0.24 & \textcolor{OliveGreen}{0.24 $\rightarrow$ 0.25} \\
Subgoal Setup ($\downarrow\uparrow$)
& \textcolor{red}{0.14 $\rightarrow$ 0.13} & \textcolor{red}{0.13 $\rightarrow$ 0.11} 
& \highlightcell{\textcolor{OliveGreen}{0.11 $\rightarrow$ 0.12}} & 0.11 $\rightarrow$ 0.11 \\
\hline
\end{tabular}
}
\vspace{2pt}
\caption{
 Demonstrating how the frequency of major reasoning pattern changes as
problem difficulty increases. 
\textcolor{OliveGreen}{Green} indicates statistically significant increases and \textcolor{red}{red} indicates significant decreases ($p < 0.05$). 
Gray boxes highlight opposing trends between OpenCodeReasoning (decrease) and \xolver (increase). 
Direction arrows indicate the expected trend direction: 
$\uparrow$ = increase, 
$\downarrow$ = decrease, 
$\downarrow\uparrow$ = mixed trend (minor decrease then elevated recovery), 
$\updownarrow$ = fluctuating trend (major decrease then recovery). While solving problems, OpenCodeReasoning struggles at Self-Evaluation and Subgoal Setup whereas \xolver overcomes it with increasing Self-Evaluation in both correct and incorrect solutions and elevated recovery in Subgoal Setup in incorrect solutions. Both OpenCodeReasoning and \xolver adapts New Approach while struggles at Problem Rephrasing.
}
\label{appendix:tab:reasoning_trace}
\end{table}

\subsection{Performance Variance Statistics}
\label{sec:appendix:variance}
In this experiment on the variance of \xolver performance, we tested \xolver against multiple runs (16 for AIME '24 and 32 for AIME '25 and LiveCodeBench) in AIME and LiveCodeBench dataset. Results shows in \Cref{tab:appendix:perf-variance} that it has small scale performance change with multiple runs which is a strong sign on the robustness of \xolver.


\begin{table}[H]
\centering
\begin{tabular}{llccc}
\hline
\textbf{Model} & \textbf{Appr.} & \textbf{AIME '24} & \textbf{AIME '25} & \textbf{LiveCodeBench (v5)} \\
\hline
o3-mini-medium & Xolver (-) & 87.2 $\pm$ 1.2 & 85.1 $\pm$ 1.3 & 79.6 $\pm$ 1.0 \\
 & Xolver (+) & 93.8 $\pm$ 0.3 & 89.4 $\pm$ 0.7 & 87.3 $\pm$ 0.4 \\
 QWQ-32B &  Xolver (-) & 89.9 $\pm$ 0.8 & 79.5 $\pm$ 1.1 & 76.2 $\pm$ 0.9 \\
 & Xolver (+) & 93.6 $\pm$ 0.2 & 82.7 $\pm$ 0.8 & 79.2 $\pm$ 0.5 \\
 o3-mini-high & Xolver (+) & 94.4 $\pm$ 0.6 & 93.7 $\pm$ 0.5 & 91.6 $\pm$ 0.3 \\
 \hline
\end{tabular}
\caption{\xolver average performance with multiple trials.}
\label{tab:appendix:perf-variance}
\end{table}

\subsection{Impact of Data-Shuffling in \xolver (+) Performance} 
\label{sec:appendix:shufflexx}
During this experiment on the impact of shuffling data on \xolver performance, we randomly shuffled the test instances and conducted the experiment with 5 runs. Results shows in \Cref{tab:appendix:data-shuffle} that \xolver has limited performance change (STD $\sim1$) with shuffling data--a strong sign on the robustness of the framework.


\begin{table}[H]
\centering
\begin{tabular}{lccccc}
\hline
\textbf{Mean $\pm$ STD } & \multicolumn{5}{c}{\textbf{(With only 5 Runs, all STD $\sim1$)} }\\
\hline
\textbf{Model} & \textbf{GSM8K} & \textbf{AIME '24} & \textbf{AIME '25} & \textbf{MATH-500} & \textbf{LiveCodeBench (v5)} \\
\hline
o3-mini-medium & 97.6 $\pm$ 1.3 & 92.2 $\pm$ 0.4 & 91.0 $\pm$ 0.3 & 98.3 $\pm$ 0.6 & 90.9 $\pm$ 1.1 \\
QWQ-32B & 97.2 $\pm$ 0.6 & 93.7 $\pm$ 0.5 & 82.7 $\pm$ 2.0 & 95.1 $\pm$ 0.6 & 83.6 $\pm$ 1.6 \\
\hline
\end{tabular}
\caption{Impact of using intermediate shared memory with shuffle of order in test set in \method{\xolver}.}
\label{tab:appendix:data-shuffle}
\end{table}

\subsection{List of Roles of Selected by Dynamic Agents} \label{sec:appendix:role-list}
\Cref{tab:appensix:list_of_roles} shows some selected specialized roles by the dynamic agents while testing on coding and math tasks along with their most frequently selected roles.
\begin{table}[h!]
\centering
\begin{tabular}{ll}
\hline
\multicolumn{2}{c}{\textbf{(a) Specialized Roles}} \\
\hline
 \textbf{Math} & \textbf{Coding} \\
\hline
Problem Analyzer & Problem Analyzer \\
Mathematical Modeler & Algorithm Designer \\
Algorithm Designer & Solution Architect \\
Numerical Analyst & Implementation Specialist \\
Symbolic Solver & Data Structure Specialist \\
Mathematician & Optimization Engineer \\
Computational Tools Specialist & Unit Tester \\
 & Debugging Expert \\
& Programmer\\
& Debugging Expert \\
& Code Reviewer \\
\hline
\multicolumn{2}{c}{\textbf{(b) Most Frequent Roles}} \\
\hline
 \textbf{Math} & \textbf{Coding} \\
 \hline
Mathematical Modeler & Algorithm Designer\\
Numerical Analyst & Implementation Specialist\\
Symbolic Solver & Data Structure Specialist\\
Mathematician & Programmer\\
Computational Tools Specialist & Optimization Engineer\\
 \hline
\end{tabular}
\caption{List of math and coding roles selected by LLMs.}
\label{tab:appensix:list_of_roles}
\end{table}




\end{document}